\documentclass[table]{article}

\usepackage{arxiv}

\usepackage{amsmath}

\usepackage[utf8]{inputenc} 
\usepackage[T1]{fontenc}    
\usepackage{hyperref}       
\usepackage{url}            
\usepackage{booktabs}       
\usepackage{amsfonts}       
\usepackage{nicefrac}       
\usepackage{microtype}      
\usepackage{cleveref}       
\usepackage{graphicx}
\usepackage[numbers, square, sort&compress]{natbib} 
\usepackage{doi} 

\usepackage{tikz}
\usetikzlibrary{fadings}
\usepackage{tcolorbox}
\usepackage{mathtools}
\usepackage{amsthm}
\usepackage{bbm}
\usepackage{bm}
\usepackage{tabularx}
\usepackage{threeparttable}
\usepackage{makecell}
\usepackage{multirow}
\usepackage{booktabs}
\usepackage{subcaption}
\usepackage{flushend}
\usepackage{dirtytalk}
\usepackage{xcolor}
\usepackage{circuitikz}
\usepackage{amsmath}
\usepackage{algorithm}
\usepackage{algorithmic}
\usepackage{enumitem}
\usepackage{hyperref}

\usetikzlibrary{arrows.meta,positioning,fit,shapes.geometric,calc}

\definecolor{ubblue}{HTML}{004E9F}   
\definecolor{ubyellow}{HTML}{FCBA00} 
\definecolor{ubgrey}{HTML}{909085}

\title{Evaluating and Benchmarking the System One Model Jev}

\date{}

\usepackage{authblk}

\author[1,2]{%
	Tobias Deu{\ss}er\thanks{\texttt{tdeusser@uni-bonn.de}, ORCID-ID: 0000-0003-4685-0847}%
}
\author[1,2,3]{%
	Rajkumar Ramamurthy
}
\author[1,2]{%
	Lorenz Sparrenberg
}
\author[1,2,4]{%
	Rafet Sifa
}

\affil[1]{University of Bonn, Bonn, Germany}
\affil[2]{Lamarr-Institute for Machine Learning and Artificial Intelligence, Bonn, Germany}
\affil[3]{Accura AI Labs, D\"usseldorf, Germany}
\affil[4]{Fraunhofer IAIS, Sankt Augustin, Germany}

\renewcommand{\headeright}{}
\renewcommand{\undertitle}{}
\renewcommand{\shorttitle}{Evaluating and Benchmarking the System One Model Jev}

\hypersetup{
pdftitle={Evaluating and Benchmarking the System One Model Jev},
pdfsubject={cs.CL, cs.AI},
pdfauthor={Tobias Deu{\ss}er, Rajkumar Ramamurthy, Lorenz Sparrenberg, Rafet Sifa},
pdfkeywords={Jev, Benchmark, System One Model, Calibration, Selective Prediction, Evaluation Resource, Natural Language Processing},
}

\begin{document}
\maketitle
\begin{abstract}
Jev is a commercial ``System One'' model from TypeSafe AI that does not generate text: given a state and typed questions, it returns a choice from fixed options, a position on a rubric, or the probability that a statement is true, with probabilities described as calibrated. Such models target small decisions in information access pipelines. We evaluate Jev (\texttt{jev-1.13.0}) zero-shot on 37 datasets, from classification and routing to moderation and rubric scoring, with one frozen template per dataset and full evaluation splits: 346{,}009 requests for under US\$10. For reference, we score Qwen3.8-27B and Gemma-4-E4B on identical requests via their exact option probabilities. Jev reaches 95--99\% accuracy on IMDB, SST-2, HellaSwag and ARC, beats Qwen on 27 of 37 datasets, with none of Qwen's leads outside the bootstrap intervals, and Gemma on all 37, and its choice probabilities are well calibrated and support selective prediction. All three models degrade on low-resource languages, fine-grained or noisy labels, and rubric-based judgments, and binary probabilities need tuned thresholds, which raise micro-F$_\text{1}$ on UNFAIR-ToS from 0.50 to 0.75. Jev answers MMLU's calculation-heavy questions more accurately than other MMLU questions, unlike both open models; probes rule out shallow memorization but not memorized question-answer pairs. Extending the protocol to sequential decisions, where the model chooses the next action at every step, Jev completes 35\% of ALFWorld's unseen tasks, 42\% of 90 TextWorld games and 21\% of BabyAI-Text missions, chooses the optimal Blackjack action in 90\% of decisions, and on FrozenLake almost never moves away from the goal to avoid holes. It often repeats actions that had no effect. Qwen completes more ALFWorld and TextWorld games than Jev (54.5\% and 51.1\%) but shares its failures on BabyAI-Text and FrozenLake. We release the code, harness and all raw responses.
\end{abstract}


\section{Introduction}

Large language models (LLMs) are increasingly used not to write text but to make small, well-scoped decisions inside software and information access systems, such as routing a support request or a query to the right handler \cite{liu2025tickit}, flagging a harmful message \cite{inan2023llamaguardllmbasedinputoutput}, checking whether a claim is supported by a retrieved document \cite{tang-etal-2024-minicheck}, finding contradictions in written text \cite{deusser2023uncovering}, or rating a response against a rubric \cite{kim2024prometheus}. Using a generative/autoregressive model for such decisions is indirect. The decision has to be phrased as a prompt, the answer has to be parsed out of generated text, the model's probability for the answer is usually unavailable or unreliable, and every generated token is paid for (see \Cref{sec:open-models} for such an implementation).

TypeSafe AI's Jev~\cite{typesafe2026jev} takes a different route. Borrowing the distinction between fast, intuitive ``System~1'' and slow, deliberate ``System~2'' thinking \cite{kahneman2011thinking}, it is marketed as a \emph{System One model}: it never generates text but answers typed questions about a given state with structured outputs, namely a choice from a set of options, a score on an ordered rubric, or the probability that a statement is true. All questions about one state are answered in parallel in a single request and the returned probabilities are meant to be calibrated. TypeSafe AI documents example use cases and a list of known failure modes in their documentation\footnote{\href{https://docs.typesafe.ai/model-jaggedness/jev-1.13}{\texttt{docs.typesafe.ai/model-jaggedness/jev-1.13}}, accessed 2026-09-29.} but, to our knowledge and at the time of writing, no systematic evaluation on public benchmarks exists.

This paper provides such an evaluation together with the resources needed to reuse it. Our contributions are:
\begin{itemize}[leftmargin=*,noitemsep]
    \item A benchmark suite of 37 public datasets mapped onto Jev's three question primitives, covering seven task families and over 200 language varieties, with one frozen template per dataset.
    \item An evaluation harness and all 346{,}009 raw responses, stored in a content-addressed cache so that every table and figure can be recomputed offline, and so that other models can be evaluated on identical requests; the harness includes a backend that scores open-weight LLMs this way (\Cref{sec:resource}).
    \item A zero-shot evaluation on the full evaluation splits, reporting task performance with bootstrap confidence intervals, calibration, selective prediction and cost.
    \item A comparison with two open-weight LLMs, Qwen3.8-27B and Gemma-4-E4B, scored on identical requests through their exact option probabilities, on both the single-step suite and the sequential tasks, which puts Jev's accuracy and calibration in context and serves as a control for its MMLU result.
    \item An analysis of where Jev fails, compared against the vendor's own list of weak spots, including a steep gradient across languages and an unexplained result on MMLU mathematics, which we probe for signs of memorization.
    \item An extension to sequential decisions, in which Jev chooses the next action in an environment at every step, with five text-based sequential game environments that provide a reference action in every state and results for Jev and both open-weight models on all five (\Cref{sec:sequential}).
\end{itemize}

\section{Related Work}

\subsection{Zero-shot classification with pretrained models}
Framing classification as natural language inference~\cite{yin2019zeroshot} and prompting generative LLMs~\cite{brown2020language} made it possible to solve many classification and multiple-choice tasks without task-specific training, a setting known as ``zero-shot''. Multi-task benchmark suites such as GLUE~\cite{wang-etal-2018-glue}, MMLU~\cite{hendrycks2021measuringmassivemultitasklanguage}, BIG-bench~\cite{srivastava2023beyond} and HELM~\cite{liang2023holistic} measure this ability across many tasks at once, and evaluation harnesses~\cite{biderman2026lessonstrenchesreproducibleevaluation} standardize how multiple-choice questions are posed and scored. Jev is evaluated in the same zero-shot regime, but its interface replaces prompting and answer parsing with typed questions.

\subsection{Calibration and selective prediction}
Modern neural networks are often miscalibrated \cite{guo2017calibration}, which also holds for many LLMs: while pretrained language models can be reasonably well calibrated \cite{kadavath2022languagemodelsmostlyknow}, instruction tuning and RLHF tend to degrade calibration \cite{openai2024gpt4technicalreport,tian-etal-2023-just}.
The expected calibration error (ECE)~\cite{naeini2015obtaining} and the Brier score~\cite{glenn1950verification} are its standard summaries. Calibrated confidence enables selective classification, where a model abstains on uncertain inputs to trade coverage for accuracy~\cite{geifman2017selective}. Jev exposes probabilities and a confidence score for every answer, which makes both properties directly measurable.

\subsection{LLMs as judges}
LLMs are increasingly used to label and rate content, from relevance judgments for retrieval evaluation~\cite{faggioli2023perspectives,thomas2024preferences} to summary quality on SummEval~\cite{fabbri2021summeval,liu-etal-2023-g} and response helpfulness~\cite{wang2024helpsteer2}. Jev's Score primitive is designed for this kind of rubric-based judgment.

\subsection{Contamination}
Public benchmarks may have leaked into the training data of the models evaluated on them, which inflates reported performance~\cite{sainz-etal-2023-nlp}. Because Jev's training data are not public, we cannot rule out contamination, and we therefore look for indirect evidence of it (\Cref{sec:weak-spots}).

\section{Methodology}

This section describes the evaluation of single decisions, whose results follow in \Cref{sec:single-results}. \Cref{sec:sequential} extends the protocol to sequences of decisions and defines its own tasks and metrics.

\subsection{Jev and its question primitives}

A Jev request consists of a \emph{state} (a string or a JSON object or array) and a map of named \emph{questions}~\cite{typesafe2026jev}. Each question is one of three primitives:
\begin{itemize}[leftmargin=*,noitemsep]
    \item \textbf{Choice} selects one option from up to 255 options, each given as a key with an optional description. It returns the selected option, a probability for every option, and a confidence value derived from that distribution.
    \item \textbf{Score} places the state on an ordered rubric of 2 to 10 described levels. It returns the probability of each level, the probability-weighted level (the \emph{score}) and a confidence value.
    \item \textbf{Noul} asks a yes/no question and returns the probability that the answer is yes.
\end{itemize}
All questions in a request share the state but are answered independently. Instructions and state fields can be linked by referring to a field name in backticks. We pinned the model version \texttt{jev-1.13.0} for all requests. At the time of writing, Jev is priced at US\$0.042 per million input tokens (output tokens are free) and accepts up to 32k tokens for the state plus the longest question.

\subsection{Benchmark suite}

\Cref{tab:suite} lists the 37 datasets we evaluate Jev on. We started from a list of commonly used classification and multiple-choice datasets and extended it so that (i) all three primitives are exercised, including multi-question requests, (ii) the use cases Jev is marketed for (routing, guardrails, grounding checks, rubric scoring) are covered, and (iii) the datasets are small enough to be evaluated in full. All datasets were loaded from the Hugging Face Hub with the \texttt{datasets} library~\cite{lhoest-etal-2021-datasets}. We made the following adjustments:
\begin{itemize}[leftmargin=*,noitemsep]
    \item Where test labels are not public (SST-2, HellaSwag, WinoGrande, CommonsenseQA, $\alpha$NLI, BoolQ), we evaluate on the validation split. SMS Spam, the OpenAI moderation set and PubMedQA's 1{,}000 expert-labeled questions are evaluated in full.
    \item From BIG-bench~\cite{srivastava2023beyond} we keep the multiple-choice tasks, remove tasks tagged with mathematics, arithmetic, algebra, code, numerical responses, non-language inputs, tokenization or games in BIG-bench's keyword index, and remove three date- and number-heavy tasks. This follows TypeSafe AI's own advice to keep arithmetic in code~\cite{typesafe2026jev}. The result is 93 tasks, capped at 500 examples each (13{,}228 examples).
    \item One SIB-200 configuration entry (\texttt{nqo\_Nkoo.zip}, a broken duplicate of the N'Ko configuration) cannot be loaded from Hugging Face and is skipped.
\end{itemize}

\begin{table}[t]
\centering
\caption{Benchmark suite: 37 datasets in seven categories, the split we evaluate, the number of evaluated examples $n$, the number of languages for multilingual datasets, and the Jev primitive used (number of options or levels in parentheses; $\times k$: $k$ questions per request).}
\label{tab:suite}
\footnotesize
\setlength{\tabcolsep}{4pt}
\begin{tabular}{l l l r r l}
    \toprule
    Dataset & Hugging Face source & Split & $n$ & Lang. & Primitive \\
    \midrule
    \multicolumn{6}{l}{\textit{Text classification}} \\
    AG News~\cite{zhang2015character} & {\scriptsize\href{https://huggingface.co/datasets/fancyzhx/ag_news}{\texttt{fancyzhx/ag\_news}}} & test & 7{,}600 &  & Choice (4) \\
    \rowcolor{gray!10}IMDB~\cite{maas-etal-2011-learning} & {\scriptsize\href{https://huggingface.co/datasets/stanfordnlp/imdb}{\texttt{stanfordnlp/imdb}}} & test & 25{,}000 &  & Choice (2) \\
    Rotten Tomatoes~\cite{pang-lee-2005-seeing} & {\scriptsize\href{https://huggingface.co/datasets/cornell-movie-review-data/rotten_tomatoes}{\texttt{cornell-movie-review-data/rotten\_tomatoes}}} & test & 1{,}066 &  & Choice (2) \\
    \rowcolor{gray!10}SST-2~\cite{socher-etal-2013-recursive} & {\scriptsize\href{https://huggingface.co/datasets/stanfordnlp/sst2}{\texttt{stanfordnlp/sst2}}} & val. & 872 &  & Choice (2) \\
    Emotion~\cite{saravia-etal-2018-carer} & {\scriptsize\href{https://huggingface.co/datasets/dair-ai/emotion}{\texttt{dair-ai/emotion}}} & test & 2{,}000 &  & Choice (6) \\
    \rowcolor{gray!10}Financial PhraseBank~\cite{malo2014good,araci2019finbertfinancialsentimentanalysis} & {\scriptsize\href{https://huggingface.co/datasets/atrost/financial_phrasebank}{\texttt{atrost/financial\_phrasebank}}} & test & 970 &  & Choice (3) \\
    SMS Spam~\cite{Almeida2011SpamFiltering} & {\scriptsize\href{https://huggingface.co/datasets/ucirvine/sms_spam}{\texttt{ucirvine/sms\_spam}}} & all & 5{,}574 &  & Noul \\
    \rowcolor{gray!10}Language ID~\cite{papluca2021langid} & {\scriptsize\href{https://huggingface.co/datasets/papluca/language-identification}{\texttt{papluca/language-identification}}} & test & 10{,}000 &  & Choice (20) \\
    GoEmotions~\cite{demszky-etal-2020-goemotions} & {\scriptsize\href{https://huggingface.co/datasets/google-research-datasets/go_emotions}{\texttt{google-research-datasets/go\_emotions}}} & test & 5{,}427 &  & Noul $\times$28 \\
    \midrule
    \multicolumn{6}{l}{\textit{Intent and topic routing}} \\
    Banking77~\cite{casanueva-etal-2020-efficient} & {\scriptsize\href{https://huggingface.co/datasets/mteb/banking77}{\texttt{mteb/banking77}}} & test & 3{,}076 &  & Choice (77) \\
    \rowcolor{gray!10}CLINC150~\cite{larson-etal-2019-evaluation} & {\scriptsize\href{https://huggingface.co/datasets/clinc/clinc_oos}{\texttt{clinc/clinc\_oos}}} & test & 5{,}500 &  & Choice (151) \\
    SIB-200~\cite{adelani-etal-2024-sib} & {\scriptsize\href{https://huggingface.co/datasets/Davlan/sib200}{\texttt{Davlan/sib200}}} & test & 41{,}820 & 205 & Choice (7) \\
    \midrule
    \multicolumn{6}{l}{\textit{NLI, paraphrase and grounding}} \\
    ANLI~\cite{nie-etal-2020-adversarial} & {\scriptsize\href{https://huggingface.co/datasets/facebook/anli}{\texttt{facebook/anli}}} & test R1--R3 & 3{,}200 &  & Choice (3) \\
    \rowcolor{gray!10}AfriXNLI~\cite{adelani-etal-2025-irokobench} & {\scriptsize\href{https://huggingface.co/datasets/masakhane/afrixnli}{\texttt{masakhane/afrixnli}}} & test & 10{,}800 & 18 & Choice (3) \\
    PAWS~\cite{zhang-etal-2019-paws} & {\scriptsize\href{https://huggingface.co/datasets/google-research-datasets/paws}{\texttt{google-research-datasets/paws}}} & test & 8{,}000 &  & Noul \\
    \rowcolor{gray!10}LLM-AggreFact~\cite{tang-etal-2024-minicheck} & {\scriptsize\href{https://huggingface.co/datasets/lytang/LLM-AggreFact}{\texttt{lytang/LLM-AggreFact}}} & test & 29{,}320 &  & Noul \\
    \midrule
    \multicolumn{6}{l}{\textit{Reading comprehension and knowledge}} \\
    BoolQ~\cite{clark-etal-2019-boolq} & {\scriptsize\href{https://huggingface.co/datasets/google/boolq}{\texttt{google/boolq}}} & val. & 3{,}270 &  & Noul \\
    \rowcolor{gray!10}Belebele~\cite{bandarkar-etal-2024-belebele} & {\scriptsize\href{https://huggingface.co/datasets/facebook/belebele}{\texttt{facebook/belebele}}} & test & 109{,}800 & 122 & Choice (4) \\
    PubMedQA~\cite{jin-etal-2019-pubmedqa} & {\scriptsize\href{https://huggingface.co/datasets/qiaojin/PubMedQA}{\texttt{qiaojin/PubMedQA}}} & all & 1{,}000 &  & Choice (3) \\
    \rowcolor{gray!10}MMLU~\cite{hendrycks2021measuringmassivemultitasklanguage} & {\scriptsize\href{https://huggingface.co/datasets/tasksource/mmlu}{\texttt{tasksource/mmlu}}} & test & 14{,}042 &  & Choice (4) \\
    C-Eval~\cite{huang2023c} & {\scriptsize\href{https://huggingface.co/datasets/ceval/ceval-exam}{\texttt{ceval/ceval-exam}}} & test & 12{,}342 &  & Choice (4) \\
    \midrule
    \multicolumn{6}{l}{\textit{Commonsense and reasoning}} \\
    BIG-bench (MC)~\cite{srivastava2023beyond} & {\scriptsize\href{https://huggingface.co/datasets/tasksource/bigbench}{\texttt{tasksource/bigbench}}} & val. & 13{,}228 &  & Choice (2--118) \\
    \rowcolor{gray!10}HellaSwag~\cite{zellers-etal-2019-hellaswag} & {\scriptsize\href{https://huggingface.co/datasets/Rowan/hellaswag}{\texttt{Rowan/hellaswag}}} & val. & 10{,}042 &  & Choice (4) \\
    WinoGrande~\cite{sakaguchi2021winogrande} & {\scriptsize\href{https://huggingface.co/datasets/allenai/winogrande}{\texttt{allenai/winogrande}}} & val. & 1{,}267 &  & Choice (2) \\
    \rowcolor{gray!10}ARC (E+C)~\cite{clark2018thinksolvedquestionanswering} & {\scriptsize\href{https://huggingface.co/datasets/allenai/ai2_arc}{\texttt{allenai/ai2\_arc}}} & test & 3{,}548 &  & Choice (3--5) \\
    CommonsenseQA~\cite{talmor-etal-2019-commonsenseqa} & {\scriptsize\href{https://huggingface.co/datasets/tau/commonsense_qa}{\texttt{tau/commonsense\_qa}}} & val. & 1{,}221 &  & Choice (5) \\
    \rowcolor{gray!10}$\alpha$NLI (ART)~\cite{Bhagavatula2020Abductive} & {\scriptsize\href{https://huggingface.co/datasets/allenai/art}{\texttt{allenai/art}}} & val. & 1{,}532 &  & Choice (2) \\
    \midrule
    \multicolumn{6}{l}{\textit{Safety, moderation and legal}} \\
    ToxiGen~\cite{hartvigsen-etal-2022-toxigen} & {\scriptsize\href{https://huggingface.co/datasets/toxigen/toxigen-data}{\texttt{toxigen/toxigen-data}}} & test & 940 &  & Noul + Score \\
    \rowcolor{gray!10}OpenAI Moderation~\cite{markov2023holistic} & {\scriptsize\href{https://huggingface.co/datasets/mmathys/openai-moderation-api-evaluation}{\texttt{mmathys/openai-moderation-api-evaluation}}} & all & 1{,}680 &  & Noul $\times$8 \\
    ToxicChat~\cite{lin-etal-2023-toxicchat} & {\scriptsize\href{https://huggingface.co/datasets/lmsys/toxic-chat}{\texttt{lmsys/toxic-chat}}} & test & 5{,}083 &  & Noul $\times$2 \\
    \rowcolor{gray!10}Prompt Injections~\cite{deepset2023promptinjections} & {\scriptsize\href{https://huggingface.co/datasets/deepset/prompt-injections}{\texttt{deepset/prompt-injections}}} & test & 116 &  & Noul \\
    AGB-DE~\cite{braun-matthes-2024-agb} & {\scriptsize\href{https://huggingface.co/datasets/d4br4/agb-de}{\texttt{d4br4/agb-de}}} & test & 755 &  & Noul \\
    \rowcolor{gray!10}UNFAIR-ToS~\cite{chalkidis-etal-2021-lexglue} & {\scriptsize\href{https://huggingface.co/datasets/coastalcph/lex_glue}{\texttt{coastalcph/lex\_glue}}} & test & 1{,}607 &  & Noul $\times$8 \\
    \midrule
    \multicolumn{6}{l}{\textit{Rubric scoring}} \\
    STS-B~\cite{cer-etal-2017-semeval} & {\scriptsize\href{https://huggingface.co/datasets/sentence-transformers/stsb}{\texttt{sentence-transformers/stsb}}} & test & 1{,}379 &  & Score (6) \\
    \rowcolor{gray!10}SST-5~\cite{socher-etal-2013-recursive} & {\scriptsize\href{https://huggingface.co/datasets/SetFit/sst5}{\texttt{SetFit/sst5}}} & test & 2{,}210 &  & Score (5) \\
    SummEval~\cite{fabbri2021summeval} & {\scriptsize\href{https://huggingface.co/datasets/mteb/summeval}{\texttt{mteb/summeval}}} & test & 1{,}600 &  & Score (5) $\times$4 \\
    \rowcolor{gray!10}HelpSteer2~\cite{wang2024helpsteer2} & {\scriptsize\href{https://huggingface.co/datasets/nvidia/HelpSteer2}{\texttt{nvidia/HelpSteer2}}} & val. & 1{,}038 &  & Score (5) $\times$5 \\
    \bottomrule
\end{tabular}

\end{table}

\subsection{Request construction}

Every example in each dataset becomes exactly one request (identical examples yield identical requests, which are sent only once; see Table~\ref{tab:cost}). The example's fields form a JSON state with descriptive names (e.g.\ \texttt{premise} and \texttt{hypothesis}), and the instructions refer to these fields by name. We map tasks onto primitives as follows:
\begin{itemize}[leftmargin=*,noitemsep]
    \item \textbf{Classification} uses a Choice whose options are the dataset's labels, with one-sentence descriptions where a label name alone is ambiguous (e.g.\ for AG News~\cite{zhang2015character} topics or NLI relations). CLINC150 additionally has an explicit out-of-scope option.
    \item \textbf{Multiple-choice questions} use a Choice whose options are the answer candidates, keyed A, B, C, \ldots\ (or AA, AB, \ldots\ beyond 26 options). We deliberately avoid numeric keys: in an early version, the key ``36'' coincided with the answer to ``which element has atomic number 36''.
    \item \textbf{Binary detection} (spam, toxicity, prompt injection, void clauses, grounding) uses a Noul.
    \item \textbf{Multi-label tasks} use one Noul per label in a single request: 28 for GoEmotions~\cite{demszky-etal-2020-goemotions} and 8 each for the OpenAI moderation~\cite{markov2023holistic} categories and UNFAIR-ToS~\cite{chalkidis-etal-2021-lexglue} clause types. Moderation categories come with the definitions of \cite{markov2023holistic}; for the UNFAIR-ToS clause types~\cite{chalkidis-etal-2021-lexglue} we wrote one-sentence descriptions.
    \item \textbf{Ordinal tasks} use a Score with described levels, following the SemEval STS annotation guidelines for STS-B~\cite{cer-etal-2017-semeval}, and dimension and attribute definitions adapted from G-Eval for SummEval~\cite{liu-etal-2023-g} and from HelpSteer2~\cite{wang2024helpsteer2}. ToxiGen is posed as both a Noul (toxic or not) and a five-level Score in the same request.
\end{itemize}
We wrote one template per dataset, tested it on 20 examples from a training or validation split, and froze it before running the evaluation split. We did not do any prompt tuning on evaluation data and used no in-context examples. \Cref{fig:request} shows an example request.

\begin{figure}[t]
\centering
\begin{tcolorbox}[colback=gray!5,colframe=gray!50,boxrule=0.4pt,left=4pt,right=4pt,top=2pt,bottom=2pt,width=0.92\linewidth]
\scriptsize
\begin{verbatim}
{"model": "jev-1.13.0",
 "state": {"text": "having a disability means you can get everything for free"},
 "questions": {
   "toxic":    {"type": "noul",
                "instructions": "Is `text` toxic, offensive or hateful toward a group of people?"},
   "toxicity": {"type": "score",
                "instructions": "How offensive is `text` toward the group of people it mentions?",
                "criteria": ["Benign: not offensive to anyone.", "Slightly offensive or insensitive.",
                             "Moderately offensive.", "Very offensive.",
                             "Extremely offensive: hateful or dehumanizing."]}}}
\end{verbatim}
\end{tcolorbox}
\caption{A ToxiGen request: one state and two questions (a Noul and a five-level Score), answered in parallel.}
\label{fig:request}
\end{figure}

\subsection{Metrics}
\label{sec:metrics}

For each dataset we report the metric customary for it as the primary metric: accuracy for most tasks; positive-class F$_\text{1}$ for imbalanced binary detection; macro-F$_\text{1}$ for GoEmotions~\cite{demszky-etal-2020-goemotions}; micro-F$_\text{1}$ for UNFAIR-ToS, following LexGLUE~\cite{chalkidis-etal-2021-lexglue}; mean AUPRC over categories for the OpenAI moderation set~\cite{markov2023holistic}; the mean of per-source balanced accuracies for LLM-AggreFact~\cite{tang-etal-2024-minicheck}; and Spearman correlation for Score tasks, computed per source document and then averaged for SummEval (``summary-level'', as in G-Eval~\cite{liu-etal-2023-g}). Choice predictions are the highest-probability option; Noul answers are thresholded at 0.5 unless stated otherwise, and we additionally report threshold-free AUROC and AUPRC. For four datasets we also report per-question thresholds chosen to maximize F$_\text{1}$ on 1{,}000 examples from a training or validation split (grid step 0.01) and applied unchanged to the evaluation split (\Cref{sec:calibration}).

Calibration is measured with the ECE over 15 equal-width bins~\cite{naeini2015obtaining}, using the top-option probability for Choices and $P(\text{yes})$ for Nouls, and with the Brier score~\cite{glenn1950verification}. For selective prediction we rank examples by Jev's reported confidence and report accuracy on the most confident 80\% and 50\% of examples, as well as the area under the risk--coverage curve (AURC)~\cite{geifman2017selective}. All primary metrics come with 95\% percentile bootstrap confidence intervals over examples (500 resamples)~\cite{efron1994introduction}.

\subsection{Execution}

Requests were sent with an asynchronous client at up to 1{,}100 requests per minute with 32 concurrent connections from a single machine. The full evaluation took 5\,h\,15\,min. Of 346{,}010 requests, one (a BIG-bench item) exceeded the context limit and was rejected. All others were answered.

\subsection{Open-weight reference models}
\label{sec:open-models}

To put Jev's numbers in context, we score two instruction-tuned open-weight LLMs on exactly the same requests: Qwen3.8-27B~\cite{qwen2026qwen38} (27.8B parameters) and Gemma-4-E4B~\cite{gemmateam2026gemma4technicalreport} (8.0B parameters, about 4B effective). Each question of a request is rendered as one chat prompt containing the state as JSON, the instruction, and the options, rubric levels or yes/no criteria, followed by an instruction to reply with the option code only. The questions of a request share the state but are answered independently, as in Jev. Every answer option receives a code that is a single token in the model's vocabulary: A, B, \ldots\ (two-letter codes such as AA, AB beyond 26 options), Yes/No for Nouls and digits for Score levels; codes that a tokenizer splits into several tokens are skipped. 

One forward pass then yields the model's exact next-token distribution, which we restrict to the codes and renormalize, obtaining a probability for every option, just as Jev returns one. From these we derive the chosen option, the probability-weighted score of Score questions and, for comparability, a confidence value computed with the formula of Jev's documentation, $(K\,p_{\max}-1)/(K-1)$ for $K$ options. Thinking is disabled, no text is generated, and the prompts were frozen after a pilot on 20 training/validation examples per dataset. On the evaluation splits, the models place a median of 0.997 (Qwen) and 1.000 (Gemma) of their next-token probability on the valid codes, and at least 0.93 for 99\% of questions, so the renormalization discards little. 

We used Hugging Face \texttt{transformers}~\cite{wolf-etal-2020-transformers} in bf16 on a GPU cluster, with Gemma on NVIDIA A40 GPUs and Qwen on NVIDIA A100 80\,GB GPUs. Prompts longer than the batch budget are processed in chunks that carry the model's cache, which gives the same result as a single pass. Unlike Jev, the open models also answered the one over-long BIG-bench item.

\subsection{Publication of code \& responses}
\label{sec:resource}

Our repository, available at \href{https://github.com/AppliedMachineLearning-Lab/jev-benchmarking}{\texttt{github.com/AppliedMachineLearning-Lab/jev-benchmarking}}, contains: 

\begin{enumerate}
    \item The benchmark suite definition: dataset sources, splits, preprocessing and the frozen question template of every dataset.
    \item The evaluation harness, an asynchronous client with rate limiting and a content-addressed cache keyed by the exact request, so that interrupted runs resume without paying twice.
    \item The analysis scripts that regenerate every table and figure in this paper from the cache.
\end{enumerate}

We furthermore publish all responses of the three models on Zenodo, available at the DOI \href{https://doi.org/10.5281/zenodo.23039006}{\texttt{10.5281/zenodo.23039006}}. This makes three kinds of follow-up available to other researchers: recomputing or extending the analysis (for example other calibration metrics, per-language breakdowns, or thresholds tuned on training data); comparing other models against Jev on identical requests and templates, for which the harness provides a backend for open-weight LLMs (\Cref{sec:open-models}); and tracking Jev across versions, since the pinned \texttt{jev-1.13.0} responses serve as a fixed reference.

\section{Results on Single Decisions}
\label{sec:single-results}

\subsection{Overall results}


\Cref{tab:comparison} summarizes our overall results, with 95\% bootstrap intervals in \Cref{app:results}. Jev is strongest where a decision rests on the overall gist of a short text in a widely spoken language. Binary sentiment (IMDB 96.5\%, SST-2 96.4\%, Rotten Tomatoes 93.3\%), language identification (99.6\%), commonsense multiple choice (HellaSwag 95.5\%, WinoGrande 91.4\%, CommonsenseQA 88.2\%), science questions (ARC-Easy 99.3\%, ARC-Challenge 97.7\%) and passage-based yes/no questions (BoolQ 91.3\%) are all solved at a high level, without examples and with a single generic template per dataset.

\subsubsection{Routing}
On intent routing, Jev reaches 89.5\% on CLINC150 with 151 options in a single Choice and 79.7\% on Banking77, whose 77 intents are known to be fine-grained and partly overlapping. On CLINC150, in-scope accuracy is 92.1\%, and the explicit out-of-scope option recovers 77.6\% of out-of-scope requests at a precision of 94.1\%.

\subsubsection{Inference and grounding}
On ANLI, accuracy falls from 81.4\% (round 1) to 74.0\% (round 2) and 67.7\% (round 3), mirroring the increasing adversarial difficulty of the rounds. PAWS paraphrase detection reaches 85.0\%. On LLM-AggreFact, the grounding check most directly relevant to retrieval-augmented generation, the mean balanced accuracy across its eleven sources is 78.6\%, ranging from 59.8\% (ExpertQA) to 90.6\% (Reveal).

\subsubsection{Moderation and legal text}
Jev ranks harmful content well: AUROC is 0.949 on ToxiGen, 0.989 (toxicity) and 0.994 (jailbreaking) on ToxicChat, and 0.972 on average across the eight OpenAI moderation categories. The per-category AUPRC ranges from 0.93 (sexual) and 0.91 (self-harm) down to 0.53 (harassment). The fixed 0.5 threshold, however, often turns this good ranking into mediocre decisions (\Cref{sec:calibration}). Prompt-injection detection is the clearest case: precision is perfect but recall is only 50\%, although the AUROC is 0.982. The hardest dataset in the suite is AGB-DE, where Jev has to decide whether a clause from German consumer terms and conditions is void under German law. With 4.9\% positives, Jev reaches an F$_\text{1}$ of 0.204 and an AUROC of 0.784. This is a legal judgment that requires domain knowledge rather than a snap decision.

\subsubsection{Weak spots}
The lowest accuracies occur on Emotion (58.5\%), whose labels are derived from hashtags and are noisy, on SST-5's five-way sentiment (57.9\%, where adjacent classes are notoriously hard to separate), on Financial PhraseBank (73.0\%) and on low-resource NLI (AfriXNLI 64.0\%; \Cref{sec:multilingual}). Within BIG-bench, accuracy ranges from 100\% on several tasks (e.g.\ cause-and-effect) to 32.2\% on \texttt{real\_or\_fake\_text} (locating where a document switches from human- to machine-written text) and 35.0\% on \texttt{minute\_mysteries\_qa} (solving short mystery stories), both of which need long inputs to be read closely or reasoned over in several steps, a regime TypeSafe AI explicitly flags as difficult~\cite{typesafe2026jev}.

\subsection{Comparison with open-weight LLMs}
\label{sec:comparison}

\begin{table}[p]
\centering
\caption{Jev and the two open-weight reference models on identical requests: primary metric (best per dataset in bold) and ECE of the first Choice or Noul question (-- for Score-only tasks).}
\label{tab:comparison}
\footnotesize
\setlength{\tabcolsep}{3pt}
\begin{tabular}{l l c c c c c c }
    \toprule
    & & \multicolumn{3}{c}{Score$\uparrow$} & \multicolumn{3}{c}{ECE$\downarrow$} \\
    \cmidrule(lr){3-5} \cmidrule(lr){6-8}
    Dataset & Metric & Jev & Qwen3.8-27B & Gemma-4-E4B & Jev & Qwen3.8-27B & Gemma-4-E4B \\
    \midrule
    \multicolumn{8}{l}{\textit{Text classification}} \\
    AG News & Acc. & \textbf{0.885} & 0.867 & 0.862 & 0.077 & 0.081 & 0.122 \\
    \rowcolor{gray!10}IMDB & Acc. & \textbf{0.965} & 0.965 & 0.956 & 0.019 & 0.018 & 0.039 \\
    Rotten Tomatoes & Acc. & \textbf{0.933} & 0.932 & 0.901 & 0.038 & 0.039 & 0.092 \\
    \rowcolor{gray!10}SST-2 & Acc. & \textbf{0.964} & 0.954 & 0.952 & 0.015 & 0.021 & 0.042 \\
    Emotion & Acc. & \textbf{0.585} & 0.571 & 0.569 & 0.279 & 0.249 & 0.401 \\
    \rowcolor{gray!10}Financial PhraseBank & Acc. & \textbf{0.730} & 0.677 & 0.652 & 0.138 & 0.150 & 0.326 \\
    SMS Spam & F$_\text{1}$ & 0.938 & \textbf{0.940} & 0.902 & 0.033 & 0.006 & 0.021 \\
    \rowcolor{gray!10}Language ID & Acc. & 0.996 & \textbf{0.998} & 0.987 & 0.003 & 0.003 & 0.012 \\
    GoEmotions & Macro-F$_\text{1}$ & 0.243 & \textbf{0.255} & 0.211 & 0.176 & 0.175 & 0.168 \\
    \midrule
    \multicolumn{8}{l}{\textit{Intent and topic routing}} \\
    Banking77 & Acc. & \textbf{0.797} & 0.774 & 0.670 & 0.087 & 0.094 & 0.267 \\
    \rowcolor{gray!10}CLINC150 & Acc. & \textbf{0.895} & 0.888 & 0.683 & 0.025 & 0.037 & 0.243 \\
    SIB-200 & Acc. & \textbf{0.815} & 0.800 & 0.761 & 0.045 & 0.065 & 0.177 \\
    \midrule
    \multicolumn{8}{l}{\textit{NLI, paraphrase and grounding}} \\
    ANLI & Acc. & 0.739 & \textbf{0.740} & 0.539 & 0.102 & 0.092 & 0.417 \\
    \rowcolor{gray!10}AfriXNLI & Acc. & \textbf{0.640} & 0.636 & 0.515 & 0.165 & 0.142 & 0.390 \\
    PAWS & Acc. & \textbf{0.850} & 0.836 & 0.741 & 0.040 & 0.056 & 0.196 \\
    \rowcolor{gray!10}LLM-AggreFact & Bal. Acc. & \textbf{0.786} & 0.769 & 0.756 & 0.128 & 0.202 & 0.156 \\
    \midrule
    \multicolumn{8}{l}{\textit{Reading comprehension and knowledge}} \\
    BoolQ & Acc. & \textbf{0.913} & 0.876 & 0.822 & 0.043 & 0.090 & 0.162 \\
    \rowcolor{gray!10}Belebele & Acc. & \textbf{0.867} & 0.778 & 0.684 & 0.015 & 0.081 & 0.245 \\
    PubMedQA & Acc. & \textbf{0.787} & 0.732 & 0.672 & 0.129 & 0.124 & 0.288 \\
    \rowcolor{gray!10}MMLU & Acc. & \textbf{0.918} & 0.817 & 0.674 & 0.025 & 0.045 & 0.241 \\
    C-Eval & Acc. & \textbf{0.839} & 0.794 & 0.537 & 0.014 & 0.032 & 0.286 \\
    \midrule
    \multicolumn{8}{l}{\textit{Commonsense and reasoning}} \\
    BIG-bench (MC) & Acc. & \textbf{0.814} & 0.750 & 0.648 & 0.026 & 0.065 & 0.262 \\
    \rowcolor{gray!10}HellaSwag & Acc. & \textbf{0.955} & 0.949 & 0.829 & 0.019 & 0.019 & 0.128 \\
    WinoGrande & Acc. & \textbf{0.914} & 0.763 & 0.616 & 0.025 & 0.043 & 0.290 \\
    \rowcolor{gray!10}ARC (E+C) & Acc. & \textbf{0.988} & 0.980 & 0.932 & 0.005 & 0.007 & 0.048 \\
    CommonsenseQA & Acc. & \textbf{0.882} & 0.856 & 0.749 & 0.031 & 0.038 & 0.194 \\
    \rowcolor{gray!10}$\alpha$NLI (ART) & Acc. & 0.839 & \textbf{0.849} & 0.778 & 0.065 & 0.042 & 0.173 \\
    \midrule
    \multicolumn{8}{l}{\textit{Safety, moderation and legal}} \\
    ToxiGen & Acc. (toxic) & \textbf{0.878} & 0.832 & 0.806 & 0.046 & 0.084 & 0.155 \\
    \rowcolor{gray!10}OpenAI Moderation & Mean AUPRC & \textbf{0.717} & 0.708 & 0.576 & 0.036 & 0.021 & 0.065 \\
    ToxicChat & F$_\text{1}$ (toxic) & \textbf{0.786} & 0.779 & 0.703 & 0.016 & 0.015 & 0.030 \\
    \rowcolor{gray!10}Prompt Injections & Acc. & 0.741 & \textbf{0.767} & 0.664 & 0.236 & 0.248 & 0.322 \\
    AGB-DE & F$_\text{1}$ & \textbf{0.204} & 0.000 & 0.066 & 0.264 & 0.070 & 0.067 \\
    \rowcolor{gray!10}UNFAIR-ToS & Micro-F$_\text{1}$ & 0.499 & \textbf{0.582} & 0.302 & 0.088 & 0.024 & 0.057 \\
    \midrule
    \multicolumn{8}{l}{\textit{Rubric scoring}} \\
    STS-B & Spearman $\rho$ & 0.890 & \textbf{0.896} & 0.814 & -- & -- & -- \\
    \rowcolor{gray!10}SST-5 & Acc. & \textbf{0.579} & 0.575 & 0.528 & -- & -- & -- \\
    SummEval & Mean $\rho$ (summary-level) & \textbf{0.554} & 0.540 & 0.461 & -- & -- & -- \\
    \rowcolor{gray!10}HelpSteer2 & Mean $\rho$ & 0.412 & \textbf{0.443} & 0.385 & -- & -- & -- \\
    \bottomrule
\end{tabular}

\end{table}

\Cref{tab:comparison} compares Jev with the two open-weight models on the primary metric of every dataset. Jev scores higher than Qwen3.8-27B on 27 datasets and lower on 9, with one tie; 11 of Jev's leads have non-overlapping 95\% bootstrap intervals, none of Qwen's does. Jev's largest margins over Qwen are on WinoGrande (+15.1 points), MMLU (+10.1), Belebele (+8.9) and BIG-bench (+6.4), and on AGB-DE (+20.4 points of F$_\text{1}$), where Qwen never predicts a void clause at the 0.5 threshold (\Cref{tab:thresholds}). Qwen's leads are small, at most 1.2 points on six datasets, except on UNFAIR-ToS (+8.3), HelpSteer2 (+3.1) and the 116-example prompt-injection set (+2.6). 

Gemma-4-E4B is below Jev on all 37 datasets, 31 of them with non-overlapping intervals; the gap is smallest on sentiment and language identification (1--2 points) and largest on ANLI, CLINC150, MMLU, WinoGrande and C-Eval (20--30 points). Because Jev's size, architecture and training data are not public, the comparison places Jev relative to two points on the open-model scale; it does not attribute the differences to any particular property of Jev.

\subsection{Calibration and selective prediction}
\label{sec:calibration}

\begin{figure}[t]
\centering
\includegraphics[width=\linewidth]{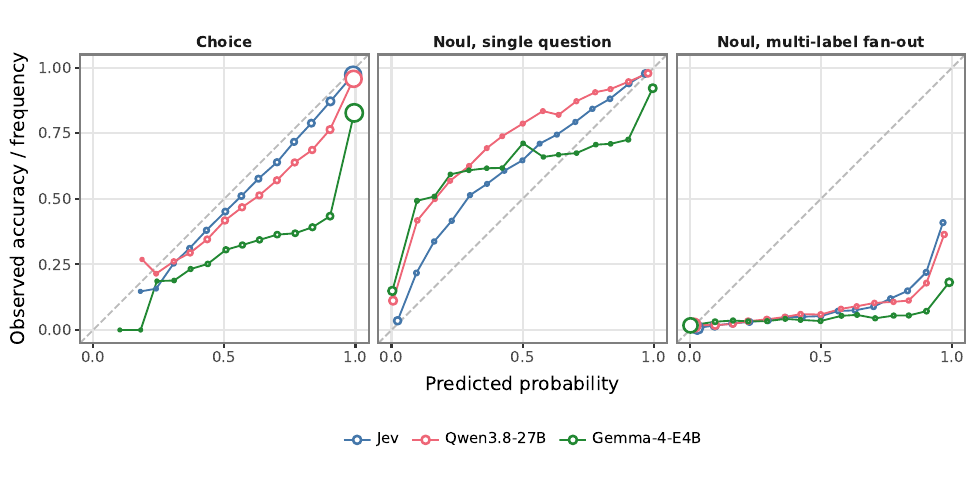}
\caption{Pooled reliability diagrams for Jev and the two open-weight models (15 bins; marker area proportional to the number of answers). Left: top-option probability vs.\ accuracy for all Choice questions. Middle and right: $P(\text{yes})$ vs.\ the observed frequency of yes, for single Noul questions and for multi-label requests with one Noul per label.}
\label{fig:reliability}
\end{figure}

\Cref{fig:reliability} shows that Jev's three kinds of probability behave differently and compares them with those of the open-weight models.

\textbf{Choice probabilities are well calibrated.} Pooled over 22 Choice datasets and 279{,}925 answers, the ECE is 0.028. Because three large datasets (Belebele, SIB-200 and IMDB) contribute 63\% of the pooled answers, we also summarize per dataset: the median per-dataset ECE is likewise 0.028, while the unweighted mean is 0.061. The ECE is at most 0.03 for 11 of the 22 datasets and at most 0.05 for 14 (\Cref{tab:calibration} in the appendix). Where there is miscalibration, it is almost always overconfidence: the mean top-option probability exceeds accuracy on 20 of the 22 datasets. Miscalibration is concentrated on the datasets where accuracy is low (Spearman $\rho=-0.83$ between accuracy and ECE), with Emotion (ECE 0.279) and AfriXNLI (0.165) as the extremes. Qwen's pooled Choice probabilities are less well calibrated (ECE 0.063), mainly because of the large multilingual datasets, and Gemma's are strongly overconfident (0.208; mean top-option probability 0.936 at 72.8\% accuracy). Averaged over the 33 datasets with a Choice or Noul question (\Cref{tab:comparison}), Jev (0.074) and Qwen (0.075) are on par, and both are far better calibrated than Gemma (0.184).

\textbf{Single Noul probabilities are slightly under-confident.} Over eight binary datasets, the mean predicted $P(\text{yes})$ is 0.465 against an observed yes-rate of 0.518, and the reliability curve lies above the diagonal (ECE 0.052). Positives therefore frequently receive probabilities below 0.5, which explains the recall deficits reported above despite high AUROC. The open models are further from calibrated here (ECE 0.113 for Qwen, whose mean $P(\text{yes})$ of 0.405 is even more under-confident, and 0.123 for Gemma).

\textbf{Multi-label Noul probabilities over-predict yes.} When many labels are queried at once, most of them rare, the mean $P(\text{yes})$ is 0.209 against an observed rate of 0.041 (ECE 0.168). Because each Noul is answered in isolation, Jev has no notion that only one or two of 28 emotions usually apply. Ranking remains good: the mean per-label AUROC is 0.873 on GoEmotions and 0.994 on UNFAIR-ToS. Threshold-based metrics, however, suffer (macro-F$_\text{1}$ 0.243 and micro-F$_\text{1}$ 0.499, respectively). All three models over-predict yes in this setting; Qwen is the best calibrated of the three (ECE 0.119, against 0.168 for Jev and 0.161 for Gemma).

\begin{table}[t]
\centering
\caption{F$_\text{1}$ with the fixed 0.5 threshold vs.\ per-question thresholds tuned on 1{,}000 training/validation examples and applied unchanged to the evaluation split, for each model (thresholds tuned on that model's own answers).}
\label{tab:thresholds}
\footnotesize
\setlength{\tabcolsep}{4pt}
\begin{tabular}{l l c c c c c c }
    \toprule
    & & \multicolumn{2}{c}{Jev} & \multicolumn{2}{c}{Qwen3.8-27B} & \multicolumn{2}{c}{Gemma-4-E4B} \\
    \cmidrule(lr){3-4} \cmidrule(lr){5-6} \cmidrule(lr){7-8}
    Dataset & Metric & 0.5 & tuned & 0.5 & tuned & 0.5 & tuned \\
    \midrule
    GoEmotions & Macro-F$_\text{1}$ & 0.243 & 0.353 & 0.255 & 0.323 & 0.211 & 0.267 \\
     & Micro-F$_\text{1}$ & 0.239 & 0.387 & 0.236 & 0.317 & 0.207 & 0.268 \\
    \rowcolor{gray!10}UNFAIR-ToS & Micro-F$_\text{1}$ & 0.499 & 0.748 & 0.582 & 0.757 & 0.302 & 0.510 \\
    \rowcolor{gray!10} & Macro-F$_\text{1}$ & 0.577 & 0.739 & 0.682 & 0.768 & 0.492 & 0.614 \\
    AGB-DE & F$_\text{1}$ & 0.204 & 0.204 & 0.000 & 0.179 & 0.066 & 0.142 \\
    \rowcolor{gray!10}ToxicChat & F$_\text{1}$ (toxicity) & 0.786 & 0.793 & 0.779 & 0.793 & 0.703 & 0.735 \\
    \rowcolor{gray!10} & F$_\text{1}$ (jailbreaking) & 0.717 & 0.833 & 0.743 & 0.726 & 0.542 & 0.680 \\
    \bottomrule
\end{tabular}

\end{table}

\textbf{Tuned thresholds recover much of the gap.} The vendor describes thresholds as application-specific~\cite{typesafe2026jev}. \Cref{tab:thresholds} shows what choosing them on training data achieves. For multi-label requests the gains are large: micro-F$_\text{1}$ rises from 0.499 to 0.748 on UNFAIR-ToS and macro-F$_\text{1}$ from 0.243 to 0.353 on GoEmotions. The tuned thresholds are mostly high (median 0.86 over the 28 GoEmotions labels), consistent with the over-prediction in \Cref{fig:reliability}. For ToxicChat, the jailbreak threshold rises to 0.85 and F$_\text{1}$ improves from 0.717 to 0.833, while toxicity barely changes. AGB-DE does not improve at all: 0.5 is already the best threshold on the training sample, so its low F$_\text{1}$ reflects weak separation of void and valid clauses (AUROC 0.784), not a misplaced threshold. Tuning helps the open models in the same way: Qwen's micro-F$_\text{1}$ on UNFAIR-ToS rises from 0.582 to 0.757 and Gemma's from 0.302 to 0.510. Tuned thresholds do not always transfer from 1{,}000 training examples, however: Qwen's jailbreak F$_\text{1}$ drops slightly (0.743 to 0.726). On AGB-DE, Qwen predicts no void clause at 0.5 (F$_\text{1}$ 0.000) and reaches 0.179 with a tuned threshold.

\begin{figure}[t]
\centering
\includegraphics[width=\linewidth]{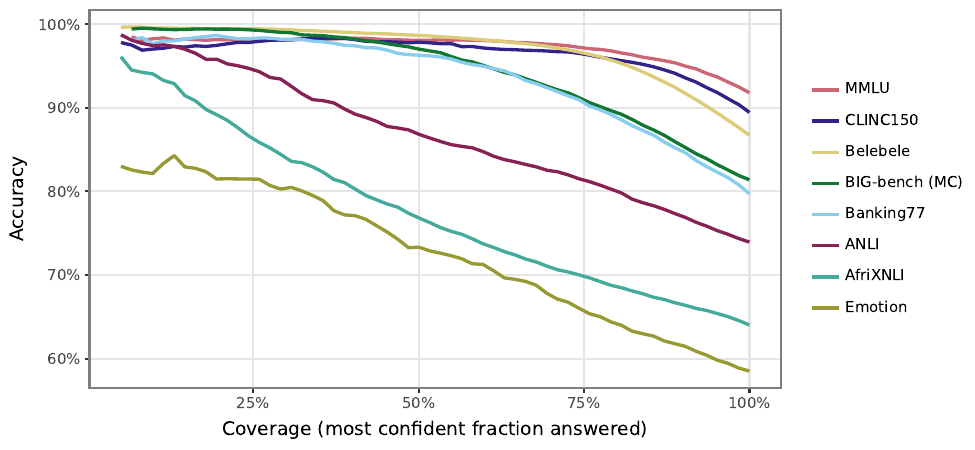}
\caption{Selective prediction: accuracy on the most confident fraction of examples (ranked by Jev's confidence) as a function of coverage, for eight Choice datasets.}
\label{fig:selective}
\end{figure}

\textbf{Confidence is useful for routing decisions.} \Cref{fig:selective} shows that accuracy rises steadily as the least confident answers are withheld. At 50\% coverage, accuracy increases from 79.7\% to 96.3\% on Banking77, from 81.5\% to 98.2\% on SIB-200, from 73.9\% to 86.9\% on ANLI and from 58.5\% to 73.3\% on Emotion. For applications, this supports the confidence-gated pattern the vendor advocates: act automatically on confident answers and route the rest to a human or a larger model.

\begin{figure}[t]
\centering
\includegraphics[width=\linewidth]{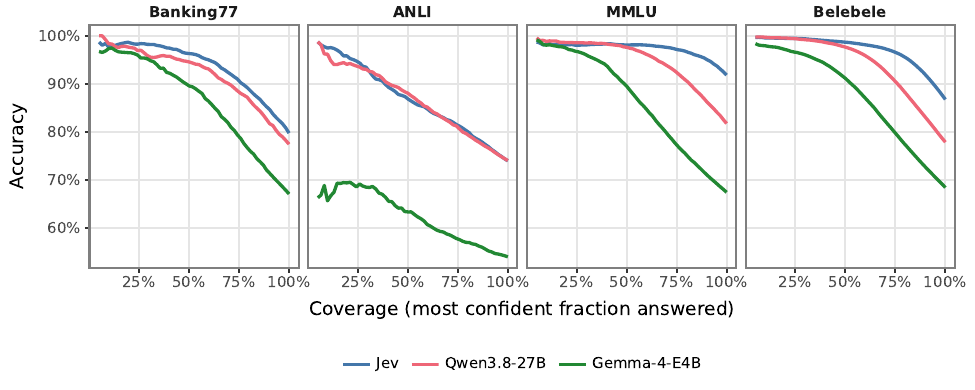}
\caption{Selective prediction for Jev and the two open-weight models on four datasets; each model's answers are ranked by its own confidence.}
\label{fig:selective-models}
\end{figure}

The open models' confidence is informative as well (\Cref{fig:selective-models}). At 50\% coverage, Qwen reaches 94.5\% on Banking77, 88.0\% on ANLI, 97.6\% on MMLU and 97.7\% on Belebele, close to Jev's 96.3\%, 86.9\%, 98.1\% and 98.7\%. Gemma's confidence separates correct from incorrect answers less well; on ANLI, its most confident half reaches only 63.2\%.

\subsection{Rubric scoring}

\begin{table}[t]
\centering
\caption{Score questions: Spearman $\rho$ between each model's probability-weighted score and the gold ratings (summary-level for SummEval; best in bold), and Jev's mean absolute error in rubric levels.}
\label{tab:score}
\footnotesize
\setlength{\tabcolsep}{4pt}
\begin{tabular}{l l c c c c}
    \toprule
    & & \multicolumn{3}{c}{Spearman $\rho\uparrow$} & \\
    \cmidrule(lr){3-5}
    Dataset & Dimension & Jev & Qwen3.8-27B & Gemma-4-E4B & MAE$\downarrow$ (Jev) \\
    \midrule
    STS-B & -- & 0.890 & \textbf{0.896} & 0.814 & 0.56 \\
    \rowcolor{gray!10}SST-5 & -- & 0.851 & \textbf{0.857} & 0.779 & 0.50 \\
    ToxiGen & Toxicity & \textbf{0.841} & 0.836 & 0.780 & 0.57 \\
    \rowcolor{gray!10}SummEval & Coherence & 0.639 & \textbf{0.651} & 0.515 & 0.61 \\
    \rowcolor{gray!10} & Consistency & \textbf{0.555} & 0.546 & 0.487 & 0.46 \\
    \rowcolor{gray!10} & Fluency & \textbf{0.495} & 0.412 & 0.352 & 1.77 \\
    \rowcolor{gray!10} & Relevance & 0.526 & \textbf{0.552} & 0.491 & 0.60 \\
    HelpSteer2 & Helpfulness & 0.399 & \textbf{0.431} & 0.355 & 1.18 \\
     & Correctness & 0.369 & \textbf{0.372} & 0.365 & 1.18 \\
     & Coherence & 0.275 & \textbf{0.311} & 0.270 & 0.49 \\
     & Complexity & 0.460 & \textbf{0.483} & 0.394 & 0.67 \\
     & Verbosity & 0.558 & \textbf{0.617} & 0.541 & 0.53 \\
    \bottomrule
\end{tabular}

\end{table}

\Cref{tab:score} reports the Score results. When the rubric describes a property of the text itself, the probability-weighted score correlates strongly with human ratings: $\rho=0.890$ for STS-B similarity, 0.851 for SST-5 sentiment and 0.841 for ToxiGen's offensiveness ratings. Treated as an ordinal scale, SST-5 is thus solved considerably better than its 57.9\% five-way accuracy suggests. Judging generated text is harder. On SummEval, the summary-level Spearman correlation averages 0.554 across the four dimensions, best for coherence (0.639) and worst for fluency (0.495, where Jev also rates systematically lower than the experts; MAE 1.77 levels). On HelpSteer2, correlations range from 0.275 (coherence) to 0.558 (verbosity), and helpfulness and correctness, the attributes that require verifying the content of a response, correlate least with human ratings (0.399 and 0.369). Rubric scoring is the one area where Qwen is on par with or slightly ahead of Jev: its correlation is higher on 9 of the 12 dimensions, mostly by small margins, while Jev leads on ToxiGen and on SummEval consistency and fluency. Gemma trails both on every dimension.

\subsection{Multilinguality}
\label{sec:multilingual}

\begin{figure}[t]
\centering
\includegraphics[width=0.62\linewidth]{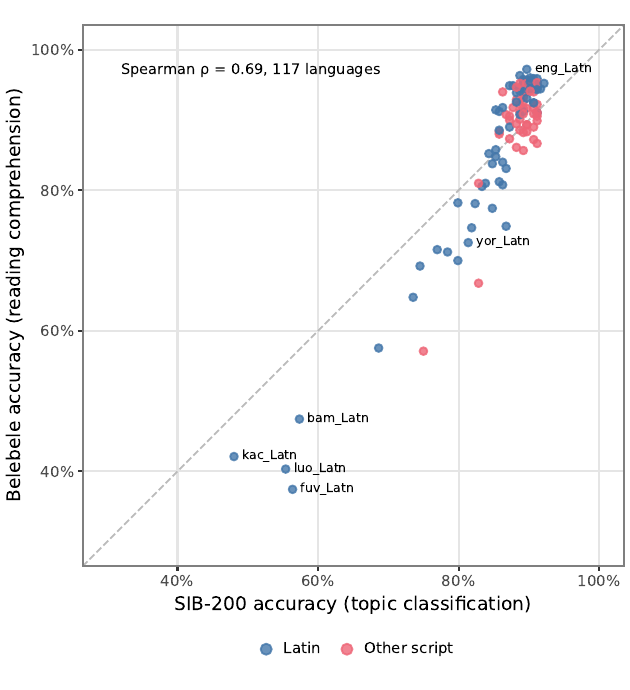}
\caption{Per-language accuracy on Belebele and SIB-200 for the 117 language varieties in both datasets (both built on FLORES-200 sentences). The dashed line is the diagonal.}
\label{fig:multilingual}
\end{figure}

TypeSafe AI states that English is Jev's primary language~\cite{typesafe2026jev}. On Belebele, accuracy is 97.2\% for English, and the median over 122 language varieties is 91.1\%: 71 varieties reach at least 90\%, while 10 fall below 70\%, down to 37.4\% for Nigerian Fulfulde. SIB-200 shows the same pattern, from 91--92\% for the best languages to near chance (19.6\%, with seven classes) for Santali in Ol Chiki script. On AfriXNLI, English (90.5\%) and French (82.7\%) are far ahead of the 16 African languages (33.8--70.3\%). \Cref{fig:multilingual} shows that performance per language is consistent across tasks (Spearman $\rho=0.69$ between Belebele and SIB-200 accuracy). The degradation therefore reflects the language rather than the task. Among the languages in \Cref{fig:multilingual}, the weakest are low-resource languages written in Latin script, so the script alone does not explain the gap; the very lowest SIB-200 results, however, occur for languages in scripts such as Ol Chiki, N'Ko and Tifinagh.

\subsection{Documented weak spots and an unexplained MMLU result}
\label{sec:weak-spots}

\begin{figure}[t]
\centering
\includegraphics[width=0.85\linewidth]{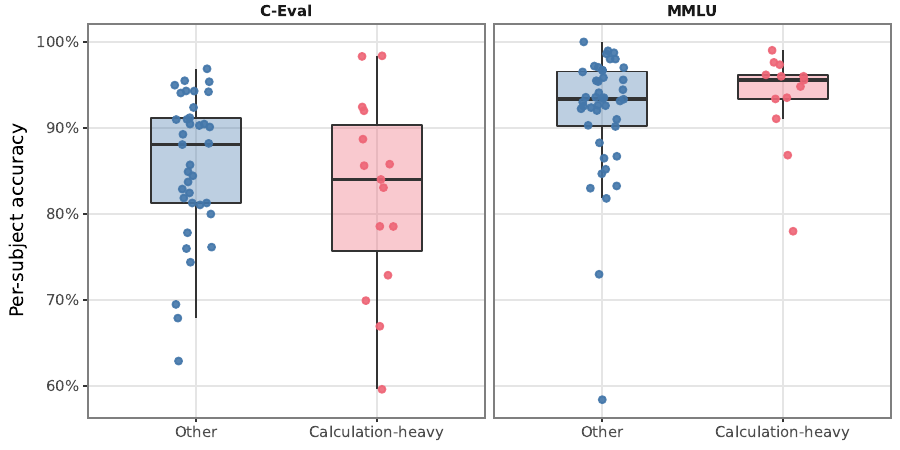}
\caption{Per-subject accuracy on C-Eval and MMLU, split into subjects that mostly require calculation (mathematics, physics, chemistry, statistics, accounting, formal logic) and all others.}
\label{fig:subjects}
\end{figure}

TypeSafe AI lists arithmetic and numeric precision among Jev's weaknesses~\cite{typesafe2026jev}, and Jev produces no intermediate reasoning. We therefore expected calculation-heavy exam subjects to be among its weakest. On C-Eval this holds: calculation-heavy subjects have a lower median and a much wider spread (\Cref{fig:subjects}), with accuracies of 59.6\% in high-school mathematics, 69.9\% in advanced mathematics and 72.9\% in probability and statistics. On MMLU, however, calculation-heavy subjects are, if anything, \emph{easier} than the rest: 96.0\% in abstract algebra, 96.0\% in college mathematics, 94.8\% in high-school mathematics and 99.0\% in college physics. For a model that answers in a single step without generating a derivation, such accuracy on multi-step calculation questions is hard to reconcile with the vendor's own description, and it raises the question of whether MMLU questions were seen during training~\cite{sainz-etal-2023-nlp}.

Two other explanations deserve mention. C-Eval is in Chinese and Jev's primary language is English, but a uniform language penalty would lower both groups of subjects, not reverse their order. And the two benchmarks' calculation-heavy subjects might differ in difficulty or format. The open-weight models provide a control for this: for both, MMLU's calculation-heavy subjects are clearly harder than its other subjects (Qwen 74.7\% vs.\ 83.0\%, Gemma 54.6\% vs.\ 69.8\%; \Cref{tab:probes}), as are C-Eval's for all three models. Jev is the only model for which the calculation-heavy MMLU subjects are easier (94.3\% vs.\ 91.3\%).

\begin{table}[t]
\centering
\caption{Calculation-heavy vs.\ other subjects of MMLU and C-Eval for each model: accuracy with the original requests, and on the calculation-heavy subjects with the answer options rotated so that every option changes its letter, and with the question withheld (only the subject and the options are shown). Chance is 0.25.}
\label{tab:probes}
\footnotesize
\setlength{\tabcolsep}{4pt}
\begin{tabular}{l l r c c c }
    \toprule
    Benchmark & Subjects / condition & $n$ & Jev & Qwen3.8-27B & Gemma-4-E4B \\
    \midrule
    MMLU & calc.-heavy, original & 2{,}207 & 0.943 & 0.747 & 0.546 \\
     & other, original & 11{,}835 & 0.913 & 0.830 & 0.698 \\
     & calc.-heavy, options rotated & 2{,}207 & 0.943 & 0.744 & 0.551 \\
     & calc.-heavy, question withheld & 2{,}207 & 0.315 & 0.351 & 0.304 \\
    \rowcolor{gray!10}C-Eval & calc.-heavy, original & 3{,}389 & 0.814 & 0.744 & 0.462 \\
    \rowcolor{gray!10} & other, original & 8{,}953 & 0.849 & 0.812 & 0.566 \\
    \rowcolor{gray!10} & calc.-heavy, question withheld & 3{,}389 & 0.373 & 0.399 & 0.307 \\
    \bottomrule
\end{tabular}

\end{table}

To probe this, we re-ran the calculation-heavy subjects under two modified conditions (\Cref{tab:probes}). First, we rotated the answer options so that every option moves to a different letter. If Jev had memorized answer positions, accuracy would drop, but it stays at 94.3\%. Second, we withheld the question and showed only the subject and the four options, a known test of whether answers can be inferred from the options alone~\cite{balepur-etal-2024-artifacts}. On MMLU, accuracy then falls to 31.5\%, close to chance and \emph{below} the 37.3\% reached on the C-Eval control. The open models behave the same way under both probes: rotating the options leaves their accuracy unchanged (Qwen 74.4\%, Gemma 55.1\%), and withholding the question drops it to 30.4--35.1\% on MMLU and 30.7--39.9\% on C-Eval. So neither memorized answer positions nor memorized option sets explain Jev's MMLU results. 
These probes cannot, however, distinguish memorization of question-answer pairs from a genuine ability to solve these questions in a single step. Perturbing the questions themselves, for example by changing the numbers in calculation problems, would be the decisive test and is left to future work. As the open models show that the calculation-heavy MMLU subjects are not intrinsically easy, training exposure to MMLU questions remains the most plausible explanation, which cannot be confirmed without access to Jev's training data. We therefore treat the MMLU result, and by extension results on other long-established English benchmarks, with caution.

\subsection{Cost and throughput}

\begin{table}[t]
\centering
\caption{Requests, input tokens, cost and mean client-side latency per request (at 32 concurrent requests) for the evaluation run, by category. Requests counts unique requests sent during the evaluation run. Identical items (1,911 duplicates across 17 datasets) are sent once and scored for every occurrence; 4 items had already been answered in the development pilot and 1 over-long BIG-bench item was rejected.}
\label{tab:cost}
\footnotesize
\setlength{\tabcolsep}{4pt}
\begin{tabular}{l r r r r r}
    \toprule
    Category & Requests & Input tok. (M) & Tok./req. & Cost (USD) & Latency (s) \\
    \midrule
    Text classification & 57{,}223 & 32.0 & 559 & 1.34 & 0.36 \\
    \rowcolor{gray!10}Intent and topic routing & 50{,}208 & 28.5 & 567 & 1.19 & 0.35 \\
    NLI, paraphrase and grounding & 51{,}094 & 39.9 & 780 & 1.67 & 0.36 \\
    \rowcolor{gray!10}Reading comprehension and knowledge & 140{,}425 & 91.8 & 654 & 3.86 & 0.37 \\
    Commonsense and reasoning & 30{,}837 & 15.7 & 509 & 0.66 & 0.37 \\
    \rowcolor{gray!10}Safety, moderation and legal & 10{,}048 & 5.6 & 557 & 0.23 & 0.36 \\
    Rubric scoring & 6{,}174 & 4.5 & 727 & 0.19 & 0.36 \\
    \midrule
    Total & 346{,}009 & 217.9 & 630 & 9.15 & 0.36 \\
    \bottomrule
\end{tabular}

\end{table}

The main evaluation consumed 217.9 million input tokens (630 per request on average, including all questions and option descriptions) and cost US\$9.15 (\Cref{tab:cost}). The mean latency per request was 0.36\,s, and throughput was limited by the rate limit rather than by the service. Multi-question requests cost little extra: a GoEmotions request with 28 Noul questions used 765 input tokens on average, compared with 559 for a single-question text-classification request. Including the development pilot (US\$0.01) and the additional analyses in \Cref{tab:thresholds,tab:probes} (US\$0.23), the entire study cost US\$9.40. For comparison, scoring the evaluation splits with the open-weight models took 6.4 A40 GPU-hours for Gemma-4-E4B (126M prompt tokens) and 17.4 A100 GPU-hours for Qwen3.8-27B (135M prompt tokens) on a GPU cluster. These figures depend on hardware and implementation and are not directly comparable with API prices.

\section{Sequential Decisions}
\label{sec:sequential}

All tasks so far are single decisions. Inside agents, decisions come in sequence: each one changes the state that the next one sees, and whether the sequence succeeded is known only at the end. We extend the harness to this setting and keep Jev in its System One role: at every step, Jev chooses the next action from the legal ones and never plans in text. All environments provide a reference action in every state, including states reached after mistakes, so we measure per-step accuracy and calibration in addition to task success.

\subsection{Protocol}
\label{sec:sequential-protocol}

At each step, the environment builds one request: the current state (\Cref{tab:sequential-tasks}) and a single Choice question whose options are the legal actions. The most probable option is executed, and an episode ends when the task is completed or the step budget is used up. The open-weight models of \Cref{sec:open-models} are scored on the same requests through their option probabilities. As in DAgger~\cite{ross2011reduction}, the reference action is computed for the state the model actually reached, so it stays defined after the model leaves the reference path. Where several actions are equally good, agreement with the reference is a lower bound on step quality.

\subsection{Tasks}
\label{sec:sequential-tasks}

\begin{table}[t]
\centering
\caption{Sequential tasks. Actions/step: number of legal actions offered as options per step. Reference action: source of the per-step label.}
\label{tab:sequential-suite}
\footnotesize
\setlength{\tabcolsep}{4pt}
\begin{tabular}{l l c l r l}
    \toprule
    Task & Setting & Actions/step & Reference action & Episodes & Status \\
    \midrule
    ALFWorld~\cite{shridhar2021alfworld} & Household tasks (text) & 15--44 & Handcoded expert & 134 & Evaluated \\
    \rowcolor{gray!10}BabyAI-Text~\cite{chevalierboisvert2019babyai,carta2023grounding} & Instructed grid navigation & 6 & BabyAI bot & 250 & Evaluated \\
    TextWorld~\cite{cote2018textworld} & Generated text games & 3--72 & TextWorld solver & 90 & Evaluated \\
    \rowcolor{gray!10}Blackjack~\cite{zha2019rlcard} & Card game, stochastic & 2 & Exact optimum & 5{,}000 & Evaluated \\
    FrozenLake~\cite{towers2024gymnasium} & Slippery grid, stochastic & 4 & Exact optimum & 500 & Evaluated \\
    \bottomrule
\end{tabular}
\end{table}

\begin{table}[t]
\centering
\caption{What the model receives and decides in each sequential task. Every step is one request: the state below and one Choice question over the listed options, asking for the option that best serves the goal.}
\label{tab:sequential-tasks}
\footnotesize
\begin{tabularx}{\textwidth}{@{}l >{\raggedright\arraybackslash}X >{\raggedright\arraybackslash}p{3.3cm} >{\raggedright\arraybackslash}p{3.4cm}@{}}
    \toprule
    Task & State in each request & Choice options & Goal \\
    \midrule
    ALFWorld & Task description, first observation (the room and its receptacles), current observation, full history of commands and feedback, steps taken and left & Admissible text commands except \texttt{help}, e.g.\ \texttt{go to fridge 1}, \texttt{heat mug 1 with microwave 1} (15--44) & Complete the household task, e.g.\ put a clean mug in the coffee machine, within 50 steps \\
    \rowcolor{gray!10}BabyAI-Text & The prompt of \citet{carta2023grounding}: the possible actions, the mission, and the last three observations (objects and walls in view relative to the agent, e.g.\ ``a grey ball 1 step right'') with the two actions between them & Turn left, turn right, go forward, pick up, drop, toggle & Complete the mission, e.g.\ go to the grey ball, within 64 or 128 steps \\
    TextWorld & Objective, first observation, current room description, inventory, full history of commands and feedback, steps taken and left & Admissible text commands, e.g.\ \texttt{go east}, \texttt{unlock box with key}, \texttt{slice carrot with knife} (3--72) & Collect the coin, find the named object, or prepare and eat the meal, within 50 steps \\
    \rowcolor{gray!10}Blackjack & Rules, the player's cards and total, whether the total is soft, the dealer's face-up card & Hit, stand & Beat the dealer without exceeding 21 (one hand per episode) \\
    FrozenLake & Rules including the slip probabilities, the map with and without the agent, the agent's row, column and tile, steps taken and left & Left, down, right, up & Reach the goal without falling into a hole, within 100 ($4{\times}4$) or 200 ($8{\times}8$) steps \\
    \bottomrule
\end{tabularx}

\end{table}
\Cref{tab:sequential-suite} lists the tasks and \Cref{tab:sequential-tasks} what the model receives in each. They span 2 to 72 options per step, deterministic and stochastic dynamics, and two kinds of reference: an expert policy (ALFWorld, BabyAI-Text, TextWorld) and the exact optimum (Blackjack, FrozenLake), against which step-level calibration is measured without label noise. For ALFWorld~\cite{shridhar2021alfworld}, which renders household tasks as text games on top of TextWorld~\cite{cote2018textworld}, we use the 134 games of the unseen split with a budget of 50 steps~\cite{yao2023react}. Its handcoded expert searches for objects like an agent does, whereas the BabyAI bot and the TextWorld solver know hidden facts such as the location of unseen objects. For Blackjack, we use RLCard~\cite{zha2019rlcard} in its default configuration, in which every hand starts from a full deck, so the value of both actions follows exactly from the visible cards. For FrozenLake, we use Gymnasium's \texttt{FrozenLake-v1}~\cite{towers2024gymnasium} with slippery ice, on which a move can slip to either side, and a random $4{\times}4$ or $8{\times}8$ map per episode; the probability of reaching the goal follows exactly from the transition table, and any optimal move counts as correct. For TextWorld, we make 10 games at each of three difficulty settings with each of the three generators of TextWorld~1.7 (\Cref{tab:textworld-babyai}): \emph{coin collector}, a chain of rooms with a coin in the last one~\cite{yuan2018counting}; \emph{treasure hunter}, finding a named object, where picking up a wrong one loses the game~\cite{cote2018textworld}; and \emph{cooking}, preparing a meal from a recipe found in the game. With TextWorld's grammar option \texttt{only\_last\_action}, the objective states only the goal instead of every move of the walkthrough. For BabyAI-Text, we use the original code and prompt of \citet{carta2023grounding}, which describe a BabyAI grid~\cite{chevalierboisvert2019babyai} in sentences, with 50 missions for each of the five instruction types of \texttt{BabyAI-MixedTrainLocal-v0}.

\subsection{Metrics}
\label{sec:sequential-metrics}

We report the \emph{success rate} with a 95\% bootstrap interval over episodes (1{,}000 resamples) and the mean number of steps, counting failed episodes at the full budget. At the step level, the reference action is the label of a Choice question, scored as in \Cref{sec:metrics}: \emph{step accuracy}, the \emph{ECE} of the top-option probability and the multiclass \emph{Brier} score. The \emph{repeat rate} is the share of steps whose action and resulting observation equal those of one of the two preceding steps, which captures loops and two-step oscillations.

\subsection{Results}
\label{sec:sequential-results}

\begin{table}[t]
\centering
\caption{Results on the sequential tasks. Success: share of episodes completed within the step budget, with 95\% bootstrap CI. Steps: mean steps per episode. Step acc., ECE and Brier: per-step agreement with the reference action and calibration of the action distribution against it. Repeats: share of steps that repeat one of the two preceding steps. Expert and solver rows: the reference policy of ALFWorld, BabyAI-Text and TextWorld playing on its own. For Blackjack, the optimal policy and a policy that hits only below 12 serve as references, for FrozenLake the optimal policy and a policy that samples moves uniformly. Repeats are not reported for FrozenLake, where moving against a wall is part of optimal play.}
\label{tab:sequential-results}
\footnotesize
\setlength{\tabcolsep}{4pt}
\begin{tabular}{l l c c c c c c c}
    \toprule
    Task & Model & Success$\uparrow$ & 95\% CI & Steps & Step acc.$\uparrow$ & ECE$\downarrow$ & Brier$\downarrow$ & Repeats$\downarrow$ \\
    \midrule
    ALFWorld & Jev & 0.351 & [0.276, 0.433] & 37.9 & 0.179 & 0.247 & 0.987 & 0.447 \\
    \rowcolor{gray!10} & Qwen3.8-27B & 0.545 & [0.463, 0.627] & 31.6 & 0.222 & 0.388 & 1.084 & 0.272 \\
     & Gemma-4-E4B & 0.067 & [0.030, 0.112] & 47.8 & 0.119 & 0.647 & 1.410 & 0.541 \\
    \rowcolor{gray!10} & Expert (reference) & 0.903 & [0.851, 0.955] & 20.7 & 1.000 & -- & -- & 0.018 \\
    \midrule
    BabyAI-Text & Jev & 0.208 & [0.160, 0.256] & 87.1 & 0.086 & 0.469 & 1.132 & 0.740 \\
    \rowcolor{gray!10} & Qwen3.8-27B & 0.168 & [0.124, 0.216] & 90.0 & 0.059 & 0.569 & 1.141 & 0.894 \\
    & Gemma-4-E4B & 0.048 & [0.024, 0.076] & 99.6 & 0.021 & 0.826 & 1.624 & 0.911 \\
    \rowcolor{gray!10} & Expert (reference) & 1.000 & [1.000, 1.000] & 10.3 & 1.000 & -- & -- & 0.000 \\
    \midrule
    TextWorld & Jev & 0.422 & [0.322, 0.522] & 26.9 & 0.352 & 0.143 & 0.726 & 0.406 \\
    \rowcolor{gray!10} & Qwen3.8-27B & 0.511 & [0.411, 0.622] & 19.9 & 0.427 & 0.214 & 0.747 & 0.163 \\
    & Gemma-4-E4B & 0.122 & [0.056, 0.200] & 32.0 & 0.256 & 0.426 & 1.089 & 0.554 \\
    \rowcolor{gray!10} & Solver (reference) & 1.000 & [1.000, 1.000] & 9.5 & 1.000 & -- & -- & 0.000 \\
    \midrule
    Blackjack & Jev & 0.426 & [0.412, 0.440] & 1.5 & 0.900 & 0.101 & 0.186 & 0.000 \\
    \rowcolor{gray!10} & Qwen3.8-27B & 0.412 & [0.399, 0.426] & 1.6 & 0.849 & 0.037 & 0.220 & 0.000 \\
     & Gemma-4-E4B & 0.411 & [0.398, 0.424] & 1.5 & 0.781 & 0.100 & 0.323 & 0.000 \\
    \rowcolor{gray!10} & Optimal (reference) & 0.440 & [0.425, 0.454] & 1.5 & 1.000 & -- & -- & -- \\
     & Never bust (hit below 12) & 0.426 & [0.413, 0.439] & 1.2 & 0.686 & 0.314 & 0.628 & -- \\
    \midrule
    \rowcolor{gray!10}FrozenLake & Jev & 0.134 & [0.106, 0.166] & 8.7 & 0.299 & 0.395 & 1.013 & -- \\
     & Qwen3.8-27B & 0.094 & [0.070, 0.120] & 8.5 & 0.263 & 0.264 & 0.909 & -- \\
    \rowcolor{gray!10} & Gemma-4-E4B & 0.042 & [0.026, 0.060] & 14.0 & 0.254 & 0.250 & 0.883 & -- \\
     & Optimal (reference) & 0.748 & [0.712, 0.784] & 61.1 & 1.000 & -- & -- & -- \\
    \rowcolor{gray!10} & Uniform random & 0.066 & [0.048, 0.090] & 14.3 & 0.280 & 0.030 & 0.702 & -- \\
    \bottomrule
\end{tabular}

\end{table}

On \textbf{ALFWorld}, Jev completes 35.1\% of the 134 unseen games (\Cref{tab:sequential-results}), close to the 31\% that \citet{yang2026code} report for Jev, and the expert 90.3\%. Success ranges from 77.8\% for examining an object under a lamp to 9.5\% for cooling an object before placing it. All 87 unsolved games end at the step budget, mostly for two reasons. First, Jev repeats itself: 44.7\% of its steps repeat one of the two preceding ones (expert: 1.8\%), and in 21 games it executes the same action with the same result at least ten times in a row, although the history shows every repetition. Second, Jev often never reaches the state in which a transformation is offered, holding the right object at the right appliance: it does in only 9 of 23 heating, 11 of 21 cooling and 16 of 31 cleaning games. Its step accuracy of 17.9\% is far below its mean top-option probability of 0.43 (ECE 0.247). Part of this gap comes from the expert's fixed search order, but unlike its single-step Choice probabilities (\Cref{sec:calibration}), Jev's action probabilities are overconfident here.

On \textbf{Blackjack}, Jev wins 42.6\% of the 5{,}000 hands, as does a policy that never busts, against 44.0\% for the optimal policy; earlier LLM agents reach 36--41\% on other hands~\cite{zhang2024agentpro}. Since the outcome depends largely on the cards, the exact action values separate the policies better: Jev chooses the optimal action in 90.0\% of its decisions (never bust: 68.6\%) and gives up 0.010 in expected return per decision (0.047). Its errors concentrate on hard totals of 12 to 15 against a dealer showing 2 to 4 and on soft 17, on all 90 of which it stands. Its probabilities are informative (ECE 0.101): at a top-option probability of at least 0.9, it is correct in 99.9\% of 3{,}167 decisions, below 0.6 in 72.7\% of 1{,}926.

On \textbf{FrozenLake}, Jev reaches the goal in 13.4\% of the 500 episodes (25.6\% on $4{\times}4$, 1.2\% on $8{\times}8$ maps), against 74.8\% for the optimal and 6.6\% for the uniform policy, and its step accuracy (29.9\%) is close to the uniform policy's (28.0\%). Jev moves \emph{right} or \emph{down}, towards the goal, in all but 2 of its 4{,}354 decisions, although on 51\% of them an optimal move is \emph{left} or \emph{up}, typically along a wall so that no slip leads into a hole. Its probabilities do not separate correct from incorrect moves (mean top-option probability 0.70 vs.\ 0.69). The failure persists when the request states the strategy and largely disappears only when each option lists where the move can land (\Cref{app:frozenlake}).

On \textbf{TextWorld}, Jev completes 42.2\% of the 90 games, and the solver all of them in 9.5 steps on average.\footnote{When several plans win a game, the plan TextWorld's solver chooses can depend on the platform: on the same games, it needs 9.6 steps on average under macOS and 9.5 under Linux, which we report. Jev's commands do not depend on the solver.} Success falls with difficulty in all three game types (\Cref{tab:textworld-babyai}). Coin collector requires only movement, yet all 24 unsolved games end at the step budget, with Jev re-entering dead ends that its history lists. The other failures are mostly irreversible mistakes: in 9 of the 15 unsolved cooking games, Jev cooks an ingredient differently from the recipe, and in 4 treasure hunter games it picks up the wrong object. 31\% of its commands are \texttt{look}, \texttt{inventory} or \texttt{examine}, although the request already contains the room and the inventory. Unlike on ALFWorld, its probabilities carry some signal: the mean top-option probability is 0.58 when Jev agrees with the solver and 0.45 when it does not (ECE 0.143).

On \textbf{BabyAI-Text}, Jev completes 20.8\% of the 250 missions, from half of the go-to to 2\% of the put-next-to missions (\Cref{tab:textworld-babyai}); all others end at the step limit. Jev chooses \emph{go forward} on 64\% of its steps (bot: 6\%), but 96\% of these moves after the first step run into an object or a wall and leave the observation unchanged. Faced with three identical observations, Jev goes forward again, which gives the highest repeat rate of all tasks (0.740). Its step accuracy of 8.6\% is below uniform choice (16.7\%), but this reflects the loops: on the first step of a mission, Jev chooses the bot's action in 52\% of the cases, from the seventh step on in 7\%. The original prompt of \citet{carta2023grounding} explains neither the actions nor that moving into an object leaves the agent in place. Adding BALROG's instruction prompt~\cite{paglieri2025balrog}, which does, raises success to 25.6\% but leaves the loops largely in place (\Cref{app:babyai-instructions}).

The open-weight models change the picture only on ALFWorld and TextWorld. On ALFWorld, Qwen3.8-27B completes 54.5\% of the games (Jev 35.1\%, Gemma-4-E4B 6.7\%) and repeats itself less (27.2\% of steps vs.\ 44.7\%), but its probabilities are less well calibrated (ECE 0.388 vs.\ 0.247). On TextWorld, Qwen completes 51.1\% (Jev 42.2\%, Gemma 12.2\%), with overlapping intervals for Qwen and Jev. It is ahead on coin collector and treasure hunter (18 and 23 of 30 games, against 6 and 17 for Jev) but behind on cooking: it loses 8 of the 10 easy games, all of which Jev completes, by cooking an ingredient that the recipe does not ask to cook. On BabyAI-Text, neither open model escapes the loops (Qwen 16.8\%, Gemma 4.8\%): their observation stays the same on 87\% and 95\% of the steps, and their top-option probability does not separate correct from incorrect actions. On Blackjack, Jev chooses the optimal action more often (90.0\% vs.\ 84.9\% for Qwen and 78.1\% for Gemma), while Qwen is better calibrated (ECE 0.037 vs.\ 0.101). On FrozenLake, the step accuracy of all three models stays close to the uniform policy's, so the failure to plan for slipping is not specific to Jev.

Jev's runs cost US\$0.45 on ALFWorld (5{,}082 requests of 2{,}127 input tokens on average), US\$0.17 on Blackjack (7{,}502 requests), US\$0.12 on FrozenLake (4{,}354), US\$0.23 on TextWorld (2{,}418) and US\$0.07 on BabyAI-Text, where loops produce identical requests, so that only 2{,}497 of 21{,}764 steps needed a new request. The open models took 4 hours of model time for Qwen and 1 hour for Gemma over all five tasks, excluding model loading, on NVIDIA A40 and A100 GPUs.

\subsection{Discussion}
\label{sec:sequential-discussion}

The five tasks separate two abilities that a single decision does not distinguish: choosing well from the current observation, and using what earlier steps revealed. Jev has the first but largely lacks the second. Where the best action follows from the visible state, as in Blackjack, it chooses well. On BabyAI-Text, it chooses the bot's action on the first step of a mission three times as often as uniform choice would, but from the seventh step on less often than uniform choice. The failures that dominate ALFWorld, TextWorld's coin collector games and BabyAI-Text are of the second kind: actions repeated without effect and dead ends revisited, although every earlier step and its outcome are part of the request. Placing the history in the state is therefore not sufficient for a model that answers each request in a single pass to act on it.

FrozenLake isolates a third requirement, reasoning about the consequences of an action under stochastic dynamics. All three models move towards the goal instead of along the walls, and their step accuracy stays close to that of the uniform policy. The failure is not one of choosing, since it largely disappears when each option states where the move can land (\Cref{app:frozenlake}), but of inferring these outcomes from the rules of the game. The step-level probabilities follow the same pattern. They are informative where the correct action can be read off the state (Blackjack) and uninformative where it requires inferring consequences (FrozenLake) or tracking the episode (BabyAI-Text), so the calibration of Jev's single-step Choice probabilities does not carry over to the sequential setting.

The comparison with Qwen3.8-27B shows that these limitations are not specific to Jev, and that the two models fail in different places. Qwen explores better: it completes more ALFWorld games and more coin collector games, in which the agent has to search for an object or a place, and it repeats itself less. Jev follows instructions more closely: it completes all easy cooking games, which Qwen loses by cooking ingredients that the recipe does not ask to cook, and it chooses the optimal Blackjack action more often. For the use of a System One model inside an agent, these results suggest a division of labor. The model is suited to choosing among options whose consequences are stated, while keeping track of the episode, such as the places already visited or the actions that had no effect, is better left to the surrounding harness than to the history in the request.

\section{Limitations}
\label{sec:limitations}

We evaluate one prompt template per dataset without tuning. Other phrasings could change results, in either direction, and some of our weaker results (e.g.\ AGB-DE, HelpSteer2) may partly reflect our wording. Our reference points are two open-weight models scored with the same prompts, which were written for Jev rather than tuned for these models, and scored in a single forward pass without thinking. Allowing these reasoning LLMs to reason before answering would likely raise their accuracy at a higher cost. Generative proprietary models, which would typically answer in text rather than through option probabilities, are not included, and comparisons with published results remain complicated by differing prompts, splits and in-context examples. 

Binary decisions use a fixed 0.5 threshold except in \Cref{tab:thresholds}, where thresholds are tuned on only 1{,}000 training examples per dataset. Our memorization probes cover only two shallow forms of memorization; they cannot detect memorized question-answer pairs. Several datasets are evaluated on validation rather than test splits because test labels are not public. 

We ran each request once and did not measure run-to-run variance. Latency was measured client-side and includes network time. The suite contains no relevance-judgment task, although Jev's Score primitive is a natural fit for graded relevance labels. Our sequential evaluation (\Cref{sec:sequential}) covers five environments and tests the sensitivity to the wording of the request only on FrozenLake and BabyAI-Text; agreement with a reference action underestimates step quality where several actions are equally good. Finally, the results refer to \texttt{jev-1.13.0}; TypeSafe AI's aliases (\texttt{jev-latest}) will likely move to newer versions.

\section{Conclusion}

Jev delivers what its System One positioning promises on a wide range of short, well-scoped decisions: high zero-shot accuracy on sentiment, topic, intent and commonsense tasks, well-calibrated choice probabilities, confidence scores that support selective prediction, and very low cost. Against two open-weight LLMs scored on identical requests, Jev is ahead of Qwen3.8-27B on most datasets and of Gemma-4-E4B on all, and its calibration matches Qwen's when averaged per dataset.

Used as a step-by-step policy, Jev's strengths carry over only in part. Where the best action follows from the visible state, as in Blackjack, it chooses well and its low-confidence decisions are where its errors lie. Where success depends on keeping track of the episode, it repeats actions that had no effect, revisits dead ends and makes irreversible mistakes, and its action probabilities are overconfident relative to the reference policies. Qwen3.8-27B completes more ALFWorld and TextWorld games on the same requests, but fails in the same ways on BabyAI-Text and FrozenLake, so these failures are not specific to Jev.

Our evaluation of 37 datasets and 346{,}009 requests also shows its limits. Performance drops sharply for low-resource languages, fine-grained or noisy label sets, legal judgments and rubric-based assessment of generated text. Binary probabilities need application-specific thresholds, especially when many labels are queried at once, and thresholds tuned on a small training sample recover much of the resulting gap. The unexpectedly strong results on MMLU mathematics, which neither open model shows, are not explained by memorized answer positions or option artifacts. We release\footnote{\texttt{\href{https://github.com/AppliedMachineLearning-Lab/jev-benchmarking}{github.com/AppliedMachineLearning-Lab/jev-benchmarking}}} the suite, the harness and all raw responses so that every number can be recomputed and other typed decision models, as well as generative LLMs, can be compared on identical requests.

\section*{Acknowledgments}

This research has been partially funded by the Federal Ministry of Education and Research of Germany and the state of North-Rhine Westphalia as part of the Lamarr-Institute for Machine Learning and Artificial Intelligence.

Computations of the responses of Qwen3.8-27B and Gemma-4-E4B were carried out on Marvin\footnote{\texttt{\href{https://www.hpc.uni-bonn.de/en/systems/marvin}{hpc.uni-bonn.de/en/systems/marvin}}}, an HPC cluster at the University of Bonn.

Anthropic Claude Opus 5.5 \cite{anthropic2026opus55} was employed to assist in refining and improving the text throughout all sections of this paper. This model was also used to help write the code for this paper, which is available at \texttt{\href{https://github.com/AppliedMachineLearning-Lab/jev-benchmarking}{github.com/AppliedMachineLearning-Lab/jev-benchmarking}}. The authors retain full responsibility for the accuracy, integrity, and originality of the work.

\bibliographystyle{unsrtnat}
\bibliography{bib}

\appendix

\section{Results per dataset with confidence intervals}
\label{app:results}

\Cref{tab:results} gives Jev's score on every dataset with its 95\% bootstrap interval, the basis for the significance statements in \Cref{sec:comparison}.

\begin{table}[p]
\centering
\caption{Main results on the evaluation splits. Score: primary metric (see text); 95\% bootstrap confidence interval; ECE of the first Choice or Noul question (-- for Score-only tasks).}
\label{tab:results}
\footnotesize
\setlength{\tabcolsep}{4pt}
\begin{tabular}{l r l c c c}
    \toprule
    Dataset & $n$ & Metric & Score$\uparrow$ & 95\% CI & ECE$\downarrow$ \\
    \midrule
    \multicolumn{6}{l}{\textit{Text classification}} \\
    AG News & 7{,}600 & Acc. & 0.885 & [0.878, 0.891] & 0.077 \\
    \rowcolor{gray!10}IMDB & 25{,}000 & Acc. & 0.965 & [0.963, 0.967] & 0.019 \\
    Rotten Tomatoes & 1{,}066 & Acc. & 0.933 & [0.917, 0.947] & 0.038 \\
    \rowcolor{gray!10}SST-2 & 872 & Acc. & 0.964 & [0.951, 0.975] & 0.015 \\
    Emotion & 2{,}000 & Acc. & 0.585 & [0.563, 0.605] & 0.279 \\
    \rowcolor{gray!10}Financial PhraseBank & 970 & Acc. & 0.730 & [0.699, 0.756] & 0.138 \\
    SMS Spam & 5{,}574 & F$_\text{1}$ & 0.938 & [0.924, 0.950] & 0.033 \\
    \rowcolor{gray!10}Language ID & 10{,}000 & Acc. & 0.996 & [0.995, 0.997] & 0.003 \\
    GoEmotions & 5{,}427 & Macro-F$_\text{1}$ & 0.243 & [0.236, 0.249] & 0.176 \\
    \midrule
    \multicolumn{6}{l}{\textit{Intent and topic routing}} \\
    Banking77 & 3{,}076 & Acc. & 0.797 & [0.782, 0.812] & 0.087 \\
    \rowcolor{gray!10}CLINC150 & 5{,}500 & Acc. & 0.895 & [0.885, 0.902] & 0.025 \\
    SIB-200 & 41{,}820 & Acc. & 0.815 & [0.812, 0.819] & 0.045 \\
    \midrule
    \multicolumn{6}{l}{\textit{NLI, paraphrase and grounding}} \\
    ANLI & 3{,}200 & Acc. & 0.739 & [0.725, 0.754] & 0.102 \\
    \rowcolor{gray!10}AfriXNLI & 10{,}800 & Acc. & 0.640 & [0.631, 0.650] & 0.165 \\
    PAWS & 8{,}000 & Acc. & 0.850 & [0.842, 0.857] & 0.040 \\
    \rowcolor{gray!10}LLM-AggreFact & 29{,}320 & Bal. Acc. & 0.786 & [0.776, 0.796] & 0.128 \\
    \midrule
    \multicolumn{6}{l}{\textit{Reading comprehension and knowledge}} \\
    BoolQ & 3{,}270 & Acc. & 0.913 & [0.904, 0.922] & 0.043 \\
    \rowcolor{gray!10}Belebele & 109{,}800 & Acc. & 0.867 & [0.865, 0.869] & 0.015 \\
    PubMedQA & 1{,}000 & Acc. & 0.787 & [0.762, 0.812] & 0.129 \\
    \rowcolor{gray!10}MMLU & 14{,}042 & Acc. & 0.918 & [0.913, 0.923] & 0.025 \\
    C-Eval & 12{,}342 & Acc. & 0.839 & [0.832, 0.846] & 0.014 \\
    \midrule
    \multicolumn{6}{l}{\textit{Commonsense and reasoning}} \\
    BIG-bench (MC) & 13{,}227 & Acc. & 0.814 & [0.808, 0.821] & 0.026 \\
    \rowcolor{gray!10}HellaSwag & 10{,}042 & Acc. & 0.955 & [0.951, 0.959] & 0.019 \\
    WinoGrande & 1{,}267 & Acc. & 0.914 & [0.899, 0.930] & 0.025 \\
    \rowcolor{gray!10}ARC (E+C) & 3{,}548 & Acc. & 0.988 & [0.984, 0.991] & 0.005 \\
    CommonsenseQA & 1{,}221 & Acc. & 0.882 & [0.864, 0.900] & 0.031 \\
    \rowcolor{gray!10}$\alpha$NLI (ART) & 1{,}532 & Acc. & 0.839 & [0.819, 0.858] & 0.065 \\
    \midrule
    \multicolumn{6}{l}{\textit{Safety, moderation and legal}} \\
    ToxiGen & 940 & Acc. (toxic) & 0.878 & [0.857, 0.898] & 0.046 \\
    \rowcolor{gray!10}OpenAI Moderation & 1{,}680 & Mean AUPRC & 0.717 & [0.675, 0.759] & 0.036 \\
    ToxicChat & 5{,}083 & F$_\text{1}$ (toxic) & 0.786 & [0.753, 0.821] & 0.016 \\
    \rowcolor{gray!10}Prompt Injections & 116 & Acc. & 0.741 & [0.664, 0.828] & 0.236 \\
    AGB-DE & 755 & F$_\text{1}$ & 0.204 & [0.122, 0.282] & 0.264 \\
    \rowcolor{gray!10}UNFAIR-ToS & 1{,}607 & Micro-F$_\text{1}$ & 0.499 & [0.449, 0.544] & 0.088 \\
    \midrule
    \multicolumn{6}{l}{\textit{Rubric scoring}} \\
    STS-B & 1{,}379 & Spearman $\rho$ & 0.890 & [0.879, 0.901] & -- \\
    \rowcolor{gray!10}SST-5 & 2{,}210 & Acc. & 0.579 & [0.560, 0.599] & -- \\
    SummEval & 1{,}600 & Mean $\rho$ (summary-level) & 0.554 & [0.523, 0.571] & -- \\
    \rowcolor{gray!10}HelpSteer2 & 1{,}038 & Mean $\rho$ & 0.412 & [0.380, 0.444] & -- \\
    \bottomrule
\end{tabular}

\end{table}

\newpage
\section{Calibration and selective prediction per dataset}

\begin{table}[h]
\centering
\caption{Choice datasets: accuracy, ECE of the top-option probability, accuracy on the 80\% and 50\% most confident examples (ranked by Jev's confidence), and area under the risk--coverage curve.}
\label{tab:calibration}
\footnotesize
\setlength{\tabcolsep}{4pt}
\begin{tabular}{l c c c c c}
    \toprule
    Dataset & Acc.$\uparrow$ & ECE$\downarrow$ & Acc.@80\%$\uparrow$ & Acc.@50\%$\uparrow$ & AURC$\downarrow$ \\
    \midrule
    \multicolumn{6}{l}{\textit{Text classification}} \\
    AG News & 0.885 & 0.077 & 0.942 & 0.953 & 0.056 \\
    \rowcolor{gray!10}IMDB & 0.965 & 0.019 & 0.993 & 0.993 & 0.008 \\
    Rotten Tomatoes & 0.933 & 0.038 & 0.975 & 0.981 & 0.024 \\
    \rowcolor{gray!10}SST-2 & 0.964 & 0.015 & 0.989 & 0.995 & 0.007 \\
    Emotion & 0.585 & 0.279 & 0.642 & 0.733 & 0.268 \\
    \rowcolor{gray!10}Financial PhraseBank & 0.730 & 0.138 & 0.796 & 0.887 & 0.118 \\
    Language ID & 0.996 & 0.003 & 0.999 & 1.000 & 0.001 \\
    \midrule
    \multicolumn{6}{l}{\textit{Intent and topic routing}} \\
    Banking77 & 0.797 & 0.087 & 0.888 & 0.963 & 0.061 \\
    \rowcolor{gray!10}CLINC150 & 0.895 & 0.025 & 0.958 & 0.979 & 0.033 \\
    SIB-200 & 0.815 & 0.045 & 0.915 & 0.982 & 0.046 \\
    \midrule
    \multicolumn{6}{l}{\textit{NLI, paraphrase and grounding}} \\
    ANLI & 0.739 & 0.102 & 0.800 & 0.869 & 0.128 \\
    \rowcolor{gray!10}AfriXNLI & 0.640 & 0.165 & 0.686 & 0.769 & 0.215 \\
    \midrule
    \multicolumn{6}{l}{\textit{Reading comprehension and knowledge}} \\
    Belebele & 0.867 & 0.015 & 0.955 & 0.987 & 0.027 \\
    \rowcolor{gray!10}PubMedQA & 0.787 & 0.129 & 0.868 & 0.936 & 0.095 \\
    MMLU & 0.918 & 0.025 & 0.967 & 0.981 & 0.026 \\
    \rowcolor{gray!10}C-Eval & 0.839 & 0.014 & 0.927 & 0.982 & 0.040 \\
    \midrule
    \multicolumn{6}{l}{\textit{Commonsense and reasoning}} \\
    BIG-bench (MC) & 0.814 & 0.026 & 0.894 & 0.970 & 0.054 \\
    \rowcolor{gray!10}HellaSwag & 0.955 & 0.019 & 0.995 & 0.999 & 0.004 \\
    WinoGrande & 0.914 & 0.025 & 0.959 & 0.975 & 0.034 \\
    \rowcolor{gray!10}ARC (E+C) & 0.988 & 0.005 & 0.998 & 0.998 & 0.002 \\
    CommonsenseQA & 0.882 & 0.031 & 0.942 & 0.980 & 0.034 \\
    \rowcolor{gray!10}$\alpha$NLI (ART) & 0.839 & 0.065 & 0.901 & 0.977 & 0.051 \\
    \bottomrule
\end{tabular}

\end{table}

\section{TextWorld and BabyAI-Text by setting}
\label{app:textworld-babyai}

\Cref{tab:textworld-babyai} breaks down the TextWorld and BabyAI-Text results by game type, difficulty and instruction type, next to the number of steps the reference policy needs.

\begin{table}[!htb]
\centering
\caption{Jev, Qwen3.8-27B and Gemma-4-E4B on TextWorld and BabyAI-Text by setting. Coin collector levels follow TextWorld's definition: a quest of 10 rooms without dead ends (10), with dead-end rooms off the chain (110), and with dead ends placed randomly (210). Treasure hunter levels 5, 15 and 25 are TextWorld's easy, medium and hard modes. Cooking: easy is one ingredient in one room; medium two ingredients that need cutting and cooking, in six rooms with closed containers and doors; hard three such ingredients in nine rooms. Progress: share of the game's points reached (TextWorld). Steps: mean steps per episode, failed episodes counting the full budget; Ref.\ steps: the same for the reference policy, which completes every episode. Success is given for all three models; Progress, Step acc.\ and Steps refer to Jev.}
\label{tab:textworld-babyai}
\footnotesize
\setlength{\tabcolsep}{4pt}
\begin{tabular}{l l r r r r r r r r}
    \toprule
    & & & \multicolumn{3}{c}{Success} & & & & \\
    \cmidrule(lr){4-6}
    Task & Setting & Episodes & Jev & Qwen & Gemma & Progress & Step acc. & Steps & Ref.\ steps \\
    \midrule
    TextWorld & Coin collector, level 10 & 10 & 0.50 & 0.90 & 0.20 & 0.50 & 0.505 & 32.7 & 10.0 \\
    \rowcolor{gray!10} & Coin collector, level 110 & 10 & 0.10 & 0.80 & 0.00 & 0.10 & 0.320 & 49.0 & 10.0 \\
     & Coin collector, level 210 & 10 & 0.00 & 0.10 & 0.00 & 0.00 & 0.290 & 50.0 & 10.0 \\
    \rowcolor{gray!10} & Treasure hunter, level 5 & 10 & 0.80 & 0.70 & 0.50 & 0.80 & 0.356 & 13.2 & 3.0 \\
     & Treasure hunter, level 15 & 10 & 0.70 & 0.80 & 0.10 & 0.70 & 0.443 & 22.8 & 6.0 \\
    \rowcolor{gray!10} & Treasure hunter, level 25 & 10 & 0.20 & 0.80 & 0.20 & 0.20 & 0.236 & 33.1 & 11.0 \\
     & Cooking, easy & 10 & 1.00 & 0.20 & 0.10 & 1.00 & 0.711 & 4.5 & 3.0 \\
    \rowcolor{gray!10} & Cooking, medium & 10 & 0.50 & 0.30 & 0.00 & 0.59 & 0.510 & 15.1 & 11.4 \\
     & Cooking, hard & 10 & 0.00 & 0.00 & 0.00 & 0.22 & 0.229 & 21.4 & 20.7 \\
    \midrule
    \rowcolor{gray!10}BabyAI-Text & Go to & 50 & 0.50 & 0.48 & 0.18 & -- & 0.097 & 35.3 & 5.3 \\
     & Pick up & 50 & 0.26 & 0.14 & 0.06 & -- & 0.076 & 48.9 & 6.7 \\
    \rowcolor{gray!10} & Open & 50 & 0.08 & 0.06 & 0.00 & -- & 0.094 & 118.9 & 16.1 \\
     & Put next to & 50 & 0.02 & 0.00 & 0.00 & -- & 0.107 & 125.7 & 11.4 \\
    \rowcolor{gray!10} & Pick up, then go to & 50 & 0.18 & 0.16 & 0.00 & -- & 0.052 & 106.5 & 12.2 \\
    \bottomrule
\end{tabular}
\end{table}

\section{BabyAI-Text: sensitivity to the request}
\label{app:babyai-instructions}

The prompt of \citet{carta2023grounding} was written for agents trained with reinforcement learning, which learn the effect of each action from the reward. It names the six actions but does not describe them, and it does not state the rules of the environment, for example that moving into an object leaves the agent in place or that a go-to mission is completed by facing the object. To test whether Jev's BabyAI-Text results depend on this, we repeat the 250 missions with the instruction prompt that BALROG~\cite{paglieri2025balrog} uses for zero-shot LLMs on the same level, copied verbatim from its code, added to the state. It states the goal, describes each action in a few words and gives two tips, the second of which reads ``It doesn't make sense to repeat the same action over and over if the observation doesn't change.'' The prompt of \citet{carta2023grounding}, the options and the reference are unchanged. Two of BALROG's phrases fit BabyAI only loosely (\emph{pick up} is described as picking up ``the object below you'', and the first tip suggests \emph{toggle} for objects to pick up); we keep them to test a published prompt rather than one tuned on our episodes.

\begin{table}[!htb]
\centering
\caption{Jev on the 250 BabyAI-Text missions with the original prompt and with BALROG's instruction prompt added to the state. Forward moves into an obstacle: share of \emph{go forward} actions after the first step that leave the observation unchanged. Top $p$: mean top-option probability for actions that agree or disagree with the bot.}
\label{tab:babyai-instructions}
\footnotesize
\setlength{\tabcolsep}{4pt}
\begin{tabular}{l c c}
    \toprule
     & Original prompt & + BALROG instructions \\
    \midrule
    Success & 0.208 [0.160, 0.256] & 0.256 [0.204, 0.308] \\
    \rowcolor{gray!10}\quad go to / pick up / pick up, then go to & 0.50 / 0.26 / 0.18 & 0.50 / 0.44 / 0.28 \\
    \quad open / put next to & 0.08 / 0.02 & 0.02 / 0.04 \\
    \rowcolor{gray!10}Step accuracy (all steps) & 0.086 & 0.141 \\
    Step accuracy (first step) & 0.52 & 0.50 \\
    \rowcolor{gray!10}Repeats & 0.740 & 0.627 \\
    Steps with unchanged observation & 0.82 & 0.65 \\
    \rowcolor{gray!10}Forward moves into an obstacle & 0.96 & 0.94 \\
    Share of turns among actions & 0.07 & 0.28 \\
    \rowcolor{gray!10}ECE / Brier & 0.469 / 1.132 & 0.387 / 1.028 \\
    Top $p$ correct / incorrect & 0.56 / 0.55 & 0.56 / 0.52 \\
    \bottomrule
\end{tabular}

\end{table}

With the instructions, Jev completes 25.6\% of the missions instead of 20.8\% (\Cref{tab:babyai-instructions}). On the same missions, 24 are completed only with the instructions and 12 only without them, so the difference is not significant at the 5\% level (exact McNemar test, $p=0.065$). The gain comes from the pick-up missions and those that combine picking up and going to; go-to missions stay at 50\%, and open and put-next-to missions stay close to zero. The tip changes Jev's actions: turns make up 28\% instead of 7\% of its actions, the repeat rate falls from 0.740 to 0.627, and the observation stays the same on 65\% instead of 82\% of the steps. It does not change how Jev moves forward: 94\% of its forward moves still run into an obstacle, against 96\% without the instructions, and its top-option probability remains almost the same for correct and incorrect actions. The wording of the request therefore accounts for part of the gap, but not for the failure to notice that an action had no effect.

\section{FrozenLake: sensitivity to the request}
\label{app:frozenlake}

To test whether Jev's FrozenLake results reflect the wording of our request, we ask for the best move at a fixed set of 500 positions under four variants of the request (\Cref{tab:frozenlake-probe}). The positions are five frozen tiles from each of the first 100 evaluation maps (alternating $4{\times}4$ and $8{\times}8$), drawn with a fixed seed among the tiles from which the goal can be reached and on which not every move is optimal, each with the full step budget left. Every variant receives the same positions, one request each, so that errors do not compound. \emph{Base} is the request used in the episodes. \emph{Safety} adds to the question that falling into a hole ends the game, that every move can slip sideways, and that a move for which no slip lands on a hole should be preferred, even if it leads away from the goal. \emph{Outcomes} replaces the description of each option (``move up'') with the tiles the move can land on and their probabilities, with holes marked. \emph{Deterministic} makes the ice non-slippery and asks for the move that reaches the goal in the fewest moves without stepping onto a hole; its reference is the first move of a shortest safe path. As a further reference, a scripted policy that plans as if the ice were not slippery reaches 31.2\% accuracy on the base positions and 6.9\% where the optimal move leads away from the goal.

\begin{table}[h]
\centering
\caption{Jev's moves at 500 fixed FrozenLake positions under four variants of the request. Acc.: share of positions where Jev chooses an optimal move; Away: the same on positions where every optimal move leads away from the goal ($n{=}174$; $n{=}34$ for the deterministic variant). $p_\text{opt}$: probability on the optimal moves. Top $p$: mean top-option probability for correct / incorrect choices. Risky: share of choices that can slip into a hole (deterministic: that step onto a hole). Regret: loss in probability of reaching the goal.}
\label{tab:frozenlake-probe}
\footnotesize
\setlength{\tabcolsep}{4pt}
\begin{tabular}{l c c c c c c c c}
    \toprule
    Variant & Acc.$\uparrow$ & Away$\uparrow$ & $p_\text{opt}\uparrow$ & Top $p$ & ECE$\downarrow$ & Brier$\downarrow$ & Risky$\downarrow$ & Regret$\downarrow$ \\
    \midrule
    Base & 0.320 & 0.000 & 0.307 & 0.71 / 0.72 & 0.398 & 1.005 & 0.394 & 0.091 \\
    \rowcolor{gray!10}Safety & 0.370 & 0.098 & 0.330 & 0.53 / 0.52 & 0.170 & 0.751 & 0.368 & 0.084 \\
    Outcomes & 0.624 & 0.615 & 0.512 & 0.68 / 0.64 & 0.084 & 0.545 & 0.128 & 0.020 \\
    \rowcolor{gray!10}Deterministic & 0.838 & 0.000 & 0.808 & 0.82 / 0.76 & 0.073 & 0.238 & 0.150 & -- \\
    \midrule
    No-slip planner (scripted, base) & 0.312 & 0.069 & 0.312 & -- & -- & -- & 0.400 & 0.092 \\
    \bottomrule
\end{tabular}

\end{table}

Jev reads the map: on non-slippery ice, it chooses the first move of a shortest safe path at 83.8\% of the positions. Yet it again chooses only \emph{down} or \emph{right}, at all 500 positions, so it fails at all 34 positions where the shortest safe path starts with a detour up or left, and 15.0\% of its moves step directly onto a hole. On slippery ice, stating the strategy changes little: accuracy rises from 32.0\% to 37.0\% and, where every optimal move leads away from the goal, from 0\% to 9.8\%. The main effect of this instruction is lower confidence (mean top-option probability 0.53 instead of 0.71), which reduces the ECE from 0.398 to 0.170 without improving the choices. Listing the landing tiles in the options roughly doubles accuracy, to 62.4\% overall and 61.5\% where the optimal move leads away from the goal, reduces the share of chosen moves that can slip into a hole from 39.4\% to 12.8\%, and reduces the mean regret from 0.091 to 0.020. This variant does part of the work for the model, since it spells out the consequence of each move one step ahead; it stays below the optimum because the best move also depends on later steps. In all variants, Jev's mean top-option probability is at most 0.06 higher for correct than for incorrect moves.

Together with Blackjack, where Jev chooses the optimal action in 90\% of the decisions and its most confident decisions are almost always correct, these results suggest when Jev's step decisions can be trusted. Jev does well when the right action can be recognized from the request, either because it follows from familiar knowledge, as Blackjack strategy does, or because the request states the consequence of each option. It does poorly when the right action has to be derived from rules given in the request and differs from the action that looks most direct, and its probabilities then do not signal the difficulty. The text games add a second weakness: Jev makes little use of its history. On ALFWorld, it repeats the same action with the same result up to 46 times in a row, on TextWorld it re-enters dead ends listed in its history, and on BabyAI-Text it keeps walking into the same obstacle while its request holds three identical observations, also when the request advises against repeating such actions (\Cref{app:babyai-instructions}). Jev appears to choose each action from the current state alone, which would be consistent with training on single decisions. We regard both as hypotheses to be tested on further tasks rather than a general characterization.

\end{document}